\documentclass[journal]{IEEEtran}
\usepackage{amsmath,amsfonts,amssymb}
\usepackage{amsthm}
\usepackage{algorithmic}
\usepackage{algorithm}
\usepackage{array}
\usepackage[caption=false,font=normalsize,labelfont=sf,textfont=sf]{subfig}
\usepackage{textcomp}
\usepackage{stfloats}
\usepackage{url}
\usepackage{verbatim}
\usepackage{graphicx}
\usepackage[sort&compress,numbers]{natbib}
\usepackage{hyperref}
\usepackage{orcidlink}
\usepackage{bm}          
\usepackage{adjustbox}   
\usepackage{placeins}    
\usepackage{multirow} 
\usepackage{booktabs} 

\newtheorem{theorem}{Theorem}[section]
\newtheorem{proposition}[theorem]{Proposition}
\newtheorem{lemma}[theorem]{Lemma}
\newtheorem{corollary}[theorem]{Corollary}
\newtheorem{canonprop}[theorem]{Proposition}
\newtheorem{remark}{Remark}

\newcommand{\sig}[2]{\ensuremath{#1_{\pm #2}}}

\def\blue #1{{\color{blue}{#1}}}
\def\red #1{{\color{red}{#1}}}

\begin{document}

\title{Sparse Hodge Preference Optimization: Fine-Tuning Language Models toward Alignment beyond Transitive Preferences}

\author{Yiming Liu~\orcidlink{0000-0002-9997-5312}, Jianwei Ma~\orcidlink{0000-0002-9803-0763},~\IEEEmembership{Senior Member,~IEEE}, Zhongwei Xing, and Tonghui Yu ~\orcidlink{0000-0002-9271-2819}
\thanks{This work was supported by the National Natural Science Foundation of China (12301340,12601571), Natural Science Foundation of Guangdong Province, China (2024A1515012761, 2026A1515011522), the HIT Research Start-up Fund. }
\thanks{Corresponding author: Tonghui Yu, Harbin Institute of Technology, Harbin, China. E-mail: tonghui\_yu@hit.edu.cn}
\thanks{Y. Liu and Z. Xing are with Department of Statistics and Data Science, Jinan University, Guangzhou, China.}
\thanks{J. Ma is with the School of Mathematics and Institute for Artificial Intelligence, Harbin Institute of Technology, Harbin, China, and the School of Earth and Space Sciences, Institute for Artificial Intelligence, Peking University, Beijing, China.}
\thanks{T. Yu is with the International Center for Interdisciplinary Statistics, School of Mathematics, Harbin Institute of Technology, Harbin, China.}
\thanks{All authors contributed equally to this work and are listed alphabetically by surname.}}

\markboth{}
{Liu \MakeLowercase{\textit{et al.}}: SHPO Fine-Tuning Language Models toward Alignment}


\maketitle

\begin{abstract}
The growing adoption of Direct Preference Optimization (DPO), which leverages pairwise preference data, reflects its great success in fine-tuning large language models to align with human expectations. It directly optimizes the policy from human preferences without explicit reward modeling or the complexities of reinforcement learning-based approaches. However, the classic DPO relies on the Bradley--Terry sampling model and cannot effectively capture intransitivity or cyclic structures in aggregated human preference data. 
In this article, we propose Sparse Hodge Preference Optimization (SHPO), a preference optimization algorithm that augments DPO with a sparse antisymmetric curl residual motivated by the combinatorial Hodge decomposition. 
We prove that SHPO strictly enlarges DPO's hypothesis class on comparison graphs containing at least one triangle, under some mild regularity
conditions on the response representations. 
We further establish a theoretical guarantee for $\ell_1$-regularized estimation of
the curl coefficients under fixed policy and representations. Qualitative studies on three preference datasets under the same training 
budget demonstrate that SHPO-aligned LLM achieves 
satisfactory performances  in preference accuracy, cyclic structure recovery, and GPT-4o-judged generation quality.
\end{abstract}

\begin{IEEEkeywords}
preference optimization, Hodge decomposition, sparse estimation, LLM alignment, cyclic preference data.
\end{IEEEkeywords}

\section{Introduction}\label{sec:introduction}
\begin{figure*}[!t]
    \centering
    \includegraphics[width=\textwidth]
    {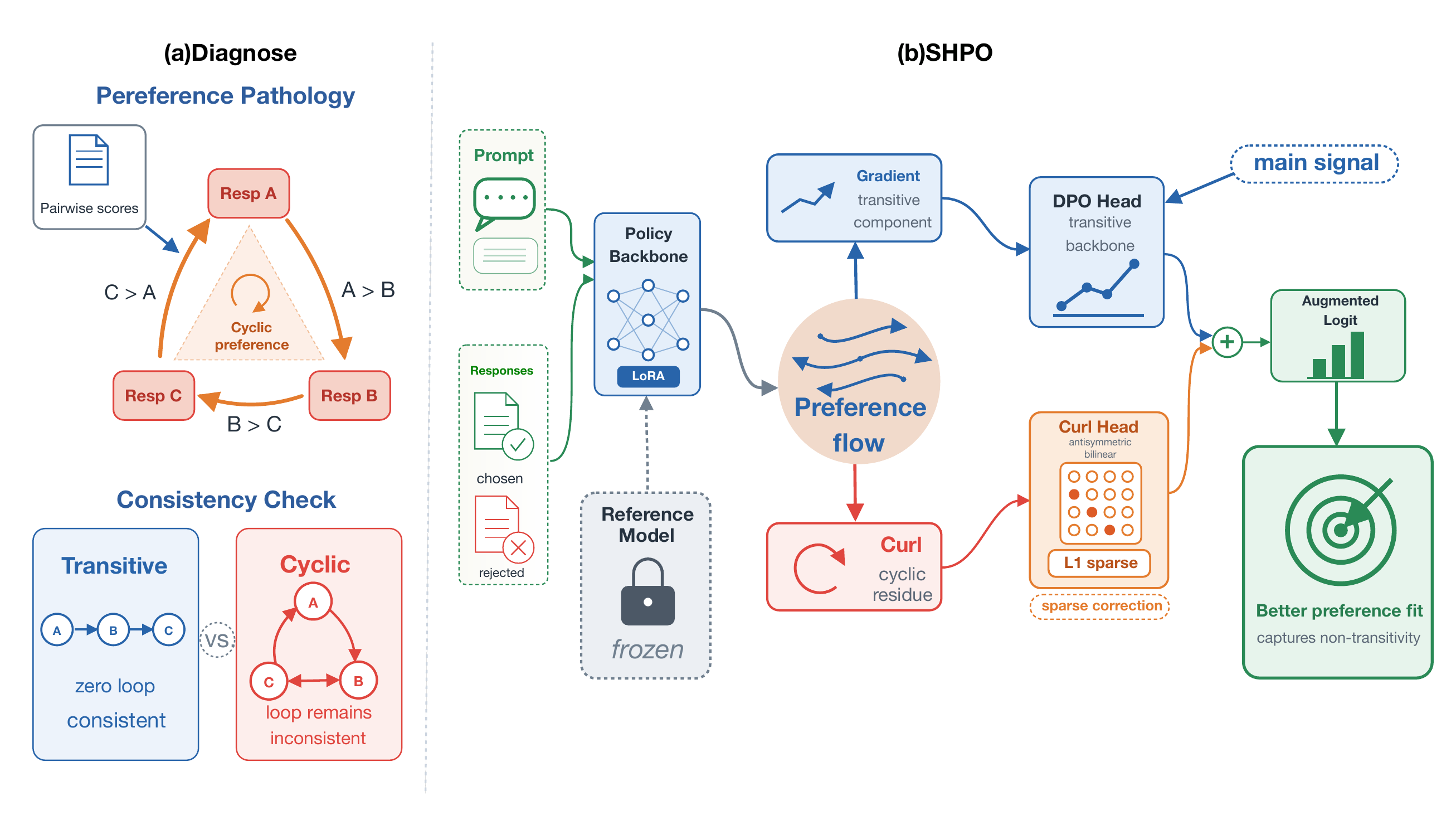}
    \caption{Outline of SHPO alignment: (a) diagnose cyclic preferences; (b)  decomposing the preference flow via Hodge into gradient, curl, and    harmonic components, and augmenting DPO with a sparse antisymmetric curl head.}
    \label{fig:overview of SHPO}
\end{figure*}

\IEEEPARstart{L}{arge} language models (LLMs) trained on human-generated  text corpora inevitably
inherit a mixture of desirable and undesirable patterns, producing outputs
that may be fluent but not always reliable or aligned with human
expectations~\cite{bai2022,ouyang2022}. Aligning such models with human
preferences is therefore critical to ensure reliable and acceptable
outputs~\cite{christiano2017}. This has motivated extensive research on
preference-based alignment, with prominent approaches including
Reinforcement Learning from Human Feedback
(RLHF)~\cite{christiano2017,ouyang2022,bai2022} and DPO~\cite{rafailov2023}. 
RLHF consists of two stages in which a reward model is first learned from human pairwise preference data, followed by policy optimization via reinforcement learning to maximize the learned reward.
Compared to RLHF, DPO integrates these two stages
into one classification objective over response pairs, simplifying the training paradigm and leading to its increasing popularity. Recent studies have
identified several critical limitations of both RLHF-based and 
DPO-based LLMs~\cite{feng2024limitations,lin2024implicit}. A typical 
concern is that transitivity underlying the reward models in RLHF and DPO is  generally not guaranteed for aggregated human
preferences~\cite{conitzer2024,liu2025}. To
illustrate this, consider three annotators comparing responses $A$, $B$, $C$
with individual rankings $A\succ B\succ C$, $B\succ C\succ A$, and
$C\succ A\succ B$. While each preference is transitive, pairwise
majority voting yields $A\succ B$, $B\succ C$, and $C\succ A$, forming
a cycle $A\succ B\succ C\succ A$ (that is, the Condorcet
paradox)~\cite{conitzer2024,mishra2023} and  making any global ranking impossible.  
Figure~\ref{fig:overview of SHPO}(a) illustrates 
this loop inconsistency
on a comparison triangle, namely, Condorcet 3-cycles. Such intransitivities often encounter 
in human judgments~\cite{tversky1969,dai2024,ge2024,liu2025}. Indeed, Liu et al. \cite{liu2025} proved that when the number of annotators is no less than three, the probability of observing Condorcet cycles approaches one at an exponential rate as the number of responses increases.  
As a result, 
both RLHF and DPO  fail to accurately represent aggregated human preferences, particularly in large-scale pairwise evaluations of language model
outputs.

This failure attributes to the scalar-reward assumption considered in both  RLHF and DPO, 
namely the Bradley-Terry (BT) model. 
It essentially assigns a scalar reward to each candidate response for a given prompt, 
with the probability of preferring one response over
another determined solely by the difference between their rewards. 
This assumption imposes a global preference ordering over all candidate responses.
As clarified in Section \ref{subsec:dpo-gradient-flow}, 
the resulting 
pairwise score is a gradient
flow and therefore has zero circulation on every comparison cycle. 
Consequently, 
it recover only the 
transitive BT projection of cyclic preference data, failing to accurately capture the underlying preference structure. 

To address these issues, numerous extensions of DPO have been proposed, which 
can be broadly categorized into two lines. 
The first line,
\emph{Within-BT methods}~\cite{azar2024,mitchell2023,chowdhury2024,meng2024,ethayarajh2024} 
improved robustness, noise tolerance, or reference-free training by modifying the loss function or training strategy.
But they still operated under the BT gradient-flow family 
thus cannot encode cycle human preferences. 
The second line, \emph{Beyond-BT methods}~\cite{zhang2025gpm,wu2024,rosset2024,munos2024} 
replaced the DPO score 
with a separately learned preference model, which is optimized through iterative self-play or online training. 
They no longer rely on the BT assumption, but leave no dedicated component to preserve the dominant transitive signal.
Both lines of existing work struggle to simultaneously model non-transitive preferences and maintain the advantages of DPO in simplicity and robustness. 

In this article, we propose SHPO, which equips DPO with the ability to accommodate cyclic preference structures through a sparse antisymmetric curl residual.
This construction is motivated by the combinatorial Hodge
decomposition~\cite{jiang2011,lim2020},  
enabling it to retain the simplicity and computational efficiency of DPO while accommodating cyclic preferences.
As demonstrated in Theorem~\ref{thm:strict-inclusion} below, 
the proposed SHPO strictly enlarges
DPO's hypothesis class on comparison graphs containing at least one triangle 
under some mild regularity conditions. 
We further provide a sparse estimation
procedure for the curl coefficients under fixed representations, 
and bounds on the rate of convergence of the  proposed $\ell_1$-penalized estimator
in Proposition~\ref{prop:sparse} below.

The remainder of this article is organized as follows. Section \ref{sec:background} presents a comparative review of related works. Sections \ref{sec:formulation}-\ref{sec:theoretical-properties} formalize the problem and present the SHPO method with its theoretical properties, respectively.  
Section \ref{sec:experiments} presents a controlled experimental evaluation of the proposed SHPO-aligned LLM on three preference datasets using a  Qwen2.5-7B-Instruct fine-tuned model, 
in comparison with its unregularized counterpart and other competing methods, together with a complementary GPT-4o judge-based assessment of generation quality.

\FloatBarrier

\section{Background and Structural Motivation}
\label{sec:background}

This section introduces the structural viewpoint used in SHPO. For a fixed prompt, pairwise preference scores can be treated as antisymmetric edge flows on a comparison graph. Under this representation, the BT model induces scalar-potential preference scores, and DPO inherits this gradient-flow structure through its reward reparameterization. This viewpoint clarifies the structural restriction of DPO-style methods and motivates the sparse curl augmentation introduced in Section~\ref{sec:formulation}.

\subsection{Preference Scores as Edge Flows}
\label{subsec:edge-flow}

Fix a prompt \(x\) and let
\[
V_x=\{y_1,\ldots,y_m\}
\]
be a finite set of candidate responses. A pairwise preference score assigns a real value \(S_{ij}(x)\) to each ordered pair \((y_i,y_j)\), where the sign of \(S_{ij}(x)\) determines the preferred direction. Since reversing the order of two responses reverses the preference direction, we impose the antisymmetry condition
\[
S_{ij}(x)=-S_{ji}(x).
\]
Thus, for each prompt, \(S(x)\) can be viewed as an antisymmetric edge flow on a comparison graph whose vertices are candidate responses and whose edges correspond to observed or admissible comparisons.

A special class of edge flows is formed by gradient flows. An edge flow \(S(x)\) is a gradient flow if there exists a scalar potential \(g_i(x)\) on each vertex such that
\[
S_{ij}(x)=g_i(x)-g_j(x).
\]
Gradient flows are path-independent, that is, any closed loop
\[
i_1 \to i_2 \to \cdots \to i_L \to i_1,
\]
their circulation is zero,
\[
S_{i_1i_2}(x)+S_{i_2i_3}(x)+\cdots+S_{i_Li_1}(x)=0.
\]
In particular, for a response triangle \((i,j,k)\),
\begin{equation}\label{eq:zero_tria_case}    
S_{ij}(x)+S_{jk}(x)+S_{ki}(x)=0.
\end{equation}
The triangular case 
with nonzero circulation indicates a local non-gradient component that cannot be represented by any scalar potential. This distinction between potential-induced flows and non-gradient flows is central to the Hodge-theoretic analysis of ranking and preference data \cite{jiang2011,lim2020}.

\subsection{DPO as a Gradient-Flow Model}
\label{subsec:dpo-gradient-flow}

The BT model represents pairwise preference probabilities through scalar reward differences \cite{bradley1952rank}, i.e.,
\[
\mathbb{P}(y_i\succ y_j\mid x)
=
\sigma\{r(x,y_i)-r(x,y_j)\},
\]
where \(r(x,y)\) is a response-level reward and \(\sigma(\cdot)\) denotes the logistic sigmoid. This scalar-reward parameterization implies that the induced pairwise score has the form of a difference between two response-level potentials.

DPO retains the BT preference model but avoids explicit reward-model training by using the KL-regularized reward reparameterization \cite{rafailov2023}
\[
r(x,y)
=
\beta \log \frac{\pi(y\mid x)}{\pi_{\mathrm{ref}}(y\mid x)}
+
\beta \log Z(x),
\]
where \(\pi\) is the trainable policy, \(\pi_{\mathrm{ref}}\) is the fixed reference policy, \(\beta>0\) is a temperature parameter, and \(Z(x)\) is a prompt-dependent normalization constant.

Since the term \(\beta\log Z(x)\) cancels in pairwise differences, the DPO score between \(y_i\) and \(y_j\) is
\[
S^{\mathrm{DPO}}_{ij}(x)
:=
\beta\left[
\log \frac{\pi(y_i\mid x)}{\pi_{\mathrm{ref}}(y_i\mid x)}
-
\log \frac{\pi(y_j\mid x)}{\pi_{\mathrm{ref}}(y_j\mid x)}
\right].
\]
Equivalently, defining
\[
g_i(x)
=
\beta\log \frac{\pi(y_i\mid x)}{\pi_{\mathrm{ref}}(y_i\mid x)},
\]
we obtain
\[
S^{\mathrm{DPO}}_{ij}(x)=g_i(x)-g_j(x).
\]
Therefore, the DPO pairwise score is a gradient flow 
and its circulation vanishes on every closed loop (see Section \ref{app:proof-dpo-gradient}). This zero-circulation property follows from the Bradley--Terry score parameterization inherited by DPO, rather than from any limitation of the underlying language-model architecture.

\subsection{Structural Gap in Existing Preference Optimization}
\label{subsec:structural-gap}

A number of recent methods improve DPO while preserving its main scalar-score structure. IPO modifies the DPO loss through a regression-style objective, while cDPO and rDPO improve robustness to noisy preference labels through conservative or regularized variants of the DPO objective \cite{azar2024,mitchell2023,chowdhury2024}. SimPO removes the explicit reference model and uses a length-normalized scalar score, and KTO replaces paired comparisons with a utility-based objective over desirable and undesirable responses \cite{meng2024,ethayarajh2024}. Although these methods differ in their loss functions and training protocols, they retain the practical advantages of DPO-style optimization, such as a simple supervised objective, stable training, and direct policy optimization without an explicit reward-modeling stage. From the edge-flow viewpoint, however, their decisions are still driven by scalar response-level quantities or scalar utilities. They therefore improve optimization and robustness within a largely transitive scoring framework, but do not directly introduce the non-gradient capacity needed to model local circulation in pairwise preferences.

Beyond-BT methods take a different direction by replacing scalar reward differences with more general pairwise or game-theoretic preference models. General pairwise preference models use more flexible comparison functions \cite{zhang2025gpm}, while self-play and Nash-style approaches formulate alignment as an equilibrium problem under general preference relations \cite{wu2024,rosset2024,munos2024}. Such methods can represent non-transitive preferences in principle, but they usually depart more substantially from the DPO score structure and may lose part of the simplicity and stability that make DPO attractive in practice. The remaining gap is therefore to introduce non-gradient preference capacity while retaining the DPO gradient component as the main transitive backbone.

\section{Formulation of SHPO}\label{sec:formulation}\label{sec:method}

The combinatorial Hodge decomposition provides a formal tool for analyzing antisymmetric edge flows on graphs and their associated clique complexes \cite{jiang2011,lim2020}. At a high level, it separates an edge flow into components associated with scalar potentials, local circulation, and global topological cycles. In preference-learning terms, these components correspond respectively to a transitive score explained by response-level potentials, local cyclic deviations, 
and global cyclic structure induced by the comparison graph.

This perspective suggests a direct design principle. Since DPO already provides a scalar-potential, gradient-flow score, a minimal structural extension should preserve this component and add only the missing local non-gradient capacity. We construct SHPO by introducing a sparse curl residual on top of the DPO logit. 
A sparse curl parameterization method is introduced to represent cyclic deviations through a small number of active curl directions. The detailed Hodge formulation and the resulting SHPO objective are given below.

\subsection{Hodge Decomposition on Preference Graphs}\label{sec:hodge-decomposition}



We assume that the zero  circulation in \eqref{eq:zero_tria_case} is violated for at least one comparison triangle. For such cycle preference structure, we apply the combinatorial Hodge decomposition  \cite{jiang2011} to orthogonally decompose the  preference flow $S_{ij}$  as gradient, curl, and harmonic components, that is, 
\begin{equation}\label{eq:hodge}
  S_{ij} = S^{\textup{grad}}_{ij} + S^{\textup{curl}}_{ij} + S^{\textup{harm}}_{ij}.
\end{equation}
Here the gradient component $S^{\textup{grad}}_{ij}$  corresponds to preference that can be explained by some scalar reward function $g\colon V \to \mathbb{R}$, namely, $S^{\textup{grad}}_{ij} = g_i - g_j$. Such preferences are necessarily transitive and admit a global ranking. 
The curl component $S^{\textup{curl}}$ captures locally cyclic preferences that cannot be explained by a global ranking and manifests as preference cycles on comparison triangles.
The harmonic component $S^{\textup{harm}}$ 
reflects residual global cyclic structure that arises from the topology of the comparison graph itself and cannot be attributed to any individual comparison triangle.



\begin{remark}\label{rem:hodge-orthogonal}
The three components
$S^{\textup{grad}}$, $S^{\textup{curl}}$, and $S^{\textup{harm}}$ are
pairwise orthogonal and the decomposition is
unique~\cite{jiang2011,lim2020}. See Appendix~\ref{app:proof-hodge} for details.
\end{remark}

Throughout this article, we restrict our attention to harmonic-free preference structures ($S^{\textup{harm}}=0$), where all cyclic inconsistencies are captured by local triangular cycles.
This condition is naturally satisfied in our setting because pairwise comparisons among the responses to each prompt form a fully connected comparison graph, leaving no residual global cyclic patterns~\cite{jiang2011}.
As a result, the preference edge flow splits orthogonally as the gradient and curl components, where the former includes the DPO pairwise score introduced in Section \ref{subsec:dpo-gradient-flow}. 


The DPO score cannot capture the curl component $S^{curl}$, so we naturally ask what an augmenting score should look like.
Such a score should satisfy three conditions.
First, it is required to be antisymmetric as clarified in Section \ref{subsec:edge-flow}.  
Second, it should lie outside the gradient-flow class. Otherwise, the model reduces to the BT family, and the structural limitation persists.
Third, it follows the fact that 
cyclic structure is usually 
weaker and more localized than the dominant transitive signal.
These requirements motivate us to consider the sparse curl parameterization below.

\subsection{Curl Reparameterization}\label{sec:sparse-curl}

For each prompt-response pair $(x,y)$, let $h(x,y)\in\mathbb{R}^{d}$ be the pooled hidden state of the response $y$ under the policy backbone. Here $d$ is the hidden dimension of the backbone. Section~\ref{sec:computational details} gives the pooling details. A bias-free linear map $W\in\mathbb{R}^{d\times 2K}$ maps $h$ to $W^{\top}h$. We then normalize this vector to unit length. The result is the curl embedding $v(x,y)\in\mathbb{R}^{2K}$. We set $v=0$ when $W^{\top}h=0$. Here $K$ is the number of curl planes, defined later in this subsection.
From the antisymmetric necessity for the curl score, we first present its canonical form in Proposition \ref{prop:canonical-skew}.
\begin{canonprop}
\label{prop:canonical-skew}
For any real antisymmetric 
matrix $A\in\mathbb{R}^{2K\times2K}$ (i.e., $A^\top=-A$),
the bilinear map
\[
  (\mathbf{v}_i,\mathbf{v}_j)\mapsto \mathbf{v}_i^\top A\,\mathbf{v}_j
  \qquad(\mathbf{v}_i,\mathbf{v}_j\in\mathbb{R}^{2K})
\]
admits an orthogonal change of basis $\tilde{\mathbf{v}} = Q\mathbf{v}$
such that
\[
  \mathbf{v}_i^\top A\,\mathbf{v}_j
  = \tilde{\mathbf{v}}_i^\top J\,\tilde{\mathbf{v}}_j,
\]
where 
$J$ is the block-diagonal matrix in the form of
\begin{equation}\label{eq:J}
  J = \operatorname{diag}\!\left(
    \begin{pmatrix} 0 & \gamma_1 \\ -\gamma_1 & 0 \end{pmatrix},\;
    \ldots,\;
    \begin{pmatrix} 0 & \gamma_K \\ -\gamma_K & 0 \end{pmatrix}
  \right).
\end{equation}
\end{canonprop}

Proposition \ref{prop:canonical-skew} follows from the Schur matrix decomposition, and its proof is detailed in Appendix~\ref{app:proof-canonical-skew}.
The orthogonal reparameterization considered in Proposition \ref{prop:canonical-skew} can be absorbed into the weight matrix of the preceding hidden layer.
Given this, the curl score between responses $y_i$
and $y_j$ is defined as 
\begin{equation}\label{eq:curl}
  c_{ij} = \mathbf{v}_i^\top J \,\mathbf{v}_j,
\end{equation}
where $J$ is given in Eq.~\eqref{eq:J}
with curl parameters $\boldsymbol{\gamma} = (\gamma_1, \ldots, \gamma_K)$.
Clearly, $c_{ij} = -c_{ji}$ (see Appendix~\ref{app:proof-antisym}), satisfying antisymmetry. By design, this score can assign nonzero triangle circulation that BT model in DPO is unable to capture. 

\begin{remark}
The score in Eq. \eqref{eq:curl} can be also expanded as
\begin{equation}\label{eq:curl_expand}
  c_{ij} = \sum_{l=1}^{K} \gamma_l
    \bigl( v_i^{(2l-1)} v_j^{(2l)} - v_i^{(2l)} v_j^{(2l-1)} \bigr),
\end{equation}
where $v_i^{(l)}$ is the $l$-th element of $\mathbf{v}_i \in \mathbb{R}^{2K}$. 
From this representation, we define $K$ curl planes, each corresponding to a two-dimensional coordinate subspace of $\mathbb{R}^{2K}$ spanned by  $(2\ell-1)$-th and $2\ell$-th coordinate axes for $\ell=1,\dots,K$. Each curl plane is also invariant under $J(\gamma)$.
\end{remark}

\begin{remark}
\label{rem:hodge-scope}
The bilinear representation 
\eqref{eq:curl} is a \emph{parametric} antisymmetric score corresponding to 
the curl mode, that is the lower-order non-gradient component in the decomposition~\eqref{eq:hodge}. It may not be exactly equal to 
the orthogonal Hodge curl subspace $S^{\textup{curl}}$ for the fixed comparison
graph, which would be 
a graph-level least-squares projection.
Recovering the exact graph-level curl component remains future
work. 
\end{remark}

\begin{remark}\label{rem:param-cost}
SHPO adds $O(d \cdot K)$ parameters to DPO's existing architecture, where $d$ is the hidden dimension and $K$ is the number of curl planes. It includes a
projection matrix $W \in \mathbb{R}^{d \times 2K}$, and a $K$-dimensional curl parameter 
$\boldsymbol{\gamma}$. Under the configuration used in our experiments ($d = 3584$, $K = 8$) in Section \ref{sec:experiments}, this amounts to $57{,}352$ additional
parameters, which is negligible compared to the base model's $O(d^2)$ parameters.
\end{remark}


\subsection{SHPO Objective}\label{sec:shpo-objective}

For responses $y_i$ and $y_j$, their joint contribution to the objective function is defined as
\begin{equation}\label{eq:logit}
  \ell_{ij}
    = \underbrace{\beta\!\left(
      \log\frac{\pi(y_i \mid x)}{\pi_{\textup{ref}}(y_i \mid x)}
      - \log\frac{\pi(y_j \mid x)}{\pi_{\textup{ref}}(y_j \mid x)}
      \right)}_{\text{gradient (DPO)}}
    + \;\alpha \, c_{ij},
\end{equation}
where $\alpha\in\mathbb{R}$ is a scalar hyperparameter controlling the curl degree and  \(\beta>0\) is a temperature parameter for DPO score. 
Write $\ell_{+-} \mathrel{:=} \ell_{ij}$ with $i = +$ (chosen) and $j = -$ (rejected),
the 
SHPO objective in population form is constructed as
\begin{equation}\label{eq:shpo}
  \mathcal{L}_{\textup{SHPO}}(\pi,W,\boldsymbol{\gamma})
    = -\,\mathbb{E}_{(x,y^+,y^-)\sim\mathcal{D}}
      \bigl[\log\sigma(\ell_{+-})\bigr]
    + \lambda_{\textup{reg}} \|\boldsymbol{\gamma}\|_1,
\end{equation}
where $\lambda_{\textup{reg}}$ is a hyperparameter controlling the sparsity of the curl planes. 
And in practice we minimize its
empirical counterpart
\begin{equation}\label{eq:shpo-emp}
  \widehat{\mathcal{L}}_{\textup{SHPO}}^{\,(n)}(\pi,W,\boldsymbol{\gamma})
    = -\frac{1}{n}\sum_{t=1}^{n} \log\sigma\bigl(\ell_{+-}^{(t)}\bigr)
      + \lambda_{\textup{reg}} \|\boldsymbol{\gamma}\|_1
\end{equation}
using the preference sample data $\mathcal{D}_n =
\{(x^{(t)}, y^{+(t)}, y^{-(t)})\}_{t=1}^{n}$  
with respect to parameters including the policy $\pi$, the projection
$W\in\mathbb{R}^{d\times 2K}$, and the curl coefficients $\boldsymbol{\gamma}=(\gamma_1,\ldots,\gamma_K)$. Among these, the matrix $W$, defined in Section~\ref{sec:sparse-curl}, produces the curl embedding $v$ used in Eq.~\eqref{eq:curl}. 

We impose $\ell_1$ penalty on $\boldsymbol{\gamma}$ into the objective function in order to remove inactive curl planes, while preserving significant curl and dominant gradient signals.
Similar sparse estimation procedures have been considered in recent work on preference learning, such as 
SparsePO \cite{christopoulou2024sparsepo}'s token-level sparse masks 
and FPO \cite{yin2024constrain}'s feature-level constraints, but differ from our SHPO's curl planes-level sparsity. 

Indeed, our proposed SHPO is close to the beyond-BT methods in motivation but differs in architecture. From the perspective of SHPO objective function, the first term in the augmented logit \eqref{eq:logit}  is the standard DPO log-ratio score, which handles the dominant transitive backbone of preferences. The second term in \eqref{eq:logit} 
captures cyclic residual structure that DPO's gradient flow cannot express. 
Rather than replacing the DPO architecture entirely, SHPO retains the policy-induced gradient term and allocates additional capacity for 
structured curl residual using an antisymmetric bilinear head whose plane coefficients are estimated via $\ell_1$-regularization. The proposed curl parameterization and augmentation strategy 
adds the minimal additional structure to the well-tested backbone, capturing cyclic deviations while preserving the gradient subspace in the original DPO structure. This technique allows it to separate the weak curl signal from the dominant gradient (transitive) signal. 

\subsection{Computational Details} \label{sec:computational details}

In Eq.\eqref{eq:shpo}-\eqref{eq:shpo-emp}, $\beta$, $\alpha$, and $\lambda_{\textup{reg}}$ are prespecified  scalar hyperparameters.
As demonstrated in Proposition~\ref{prop:sparse}, with the choice  $\lambda_{\textup{reg}}\asymp|\alpha|L\sqrt{\log K / n}$, 
we can bound the estimation loss with high probability. And in numeric examples, we consider $\lambda_{\textup{reg}}=4|\alpha|L\sqrt{\log K/n}$.

We note that $\alpha$ is theoretically unidentifiable, as $(\alpha,\boldsymbol{\gamma})$ and $(1,\alpha\boldsymbol{\gamma})$ induces the same value of  Eq.\eqref{eq:logit}. Nevertheless, retaining $\alpha$ offers practical advantages for numerical optimization. On the one hand, $\alpha$ scales the gradient  of the curl head relative to the DPO term, thereby controlling their effective learning rate. On the other hand, under an $\ell_1$ regularization, this scaling balances the data-fitting term and the sparsity penalty, effectively modulating the regularization strength.
Thus, we tune  $\alpha$ as a  hyperparameter  in our experiments (see Section \ref{sec:experimental-setup}), but fix $\alpha=1$ in theoretical analysis (Section \ref{sec:theoretical-properties}). 

Once these hyperparameters have been determined, 
the SHPO model is trained end-to-end as follows. 
First, a mini-batch of preference triples $(x,y_+,y_-)$, drawn from the training preference dataset $\mathcal{D}_n=\{(x^{(t)},y_+^{(t)},y_-^{(t)})\}_{t=1}^{n}$, is fed into the policy backbone. Second, the log-ratios $\log\{\pi(y_{\pm}\mid x)/\pi_{\mathrm{ref}}(y_{\pm}\mid x)\}$ and the gradient (DPO) term in \eqref{eq:logit} are then evaluated by the policy and reference models.  Third, the hidden representation $h_+,h_-$, corresponding to the chosen and rejected responses respectively, are obtained by mean-pooling the last-layer hidden states over response tokens, and are subsequently mapped by a bias-free linear projection $W\in\mathbb{R}^{d\times 2K}$ to $\mathbb{R}^{2K}$, followed by $\ell_2$ normalization to produce the curl embeddings $v_+,v_-$ and the curl score $c_{+-}$ as in Eq. \eqref{eq:curl}. 
Finally, the empirical SHPO loss \eqref{eq:shpo-emp}  is calculated by averaging over the mini-batch, gradients are backpropagated, and all trainable parameters $(\pi,W,\gamma)$ are  updated using a stochastic gradient optimizer. The $\ell_1$ regularization on $\gamma$ is incorporated into this gradient computation through its subgradient.  

We remark here that since the curl head shares the backbone representation with the DPO term, gradients derived from both components also flow through the shared backbone parameters. We will examine this coupling empirically by the detach ablation in Section \ref{sec:ablation-studies}. In addition,  Qwen 2.5-7B-Instruct base model is considered in our experiments in Section \ref{sec:experiments}, and its pre-trained checkpoint serves as the reference $\pi_{\mathrm{ref}}$ which remains frozen throughout training and requires only forward evaluations. The same procedure can be applied to other base and reference models. 
 

\section{Theoretical Properties}
\label{sec:theoretical-properties}

We study the theoretical properties of SHPO under the preference-flow view introduced in Sections~\ref{sec:background} and~\ref{sec:method}. The DPO term gives a gradient-flow score, while the curl head adds a non-gradient correction. We first show that DPO is obtained as a special case of SHPO. We then prove that the inclusion is strict when the comparison graph contains a nondegenerate response triangle. Finally, we give a conditional sparse-estimation result for the curl coefficients when the policy and response representations are fixed.

\begin{corollary}\label{cor:dpo-special}
Let $W$ be any projection matrix used to form the curl embeddings in
Section~III-B. If $\gamma = 0$, then the curl score in (3) is
identically zero and
\[
\mathcal{L}_{\mathrm{SHPO}}(\pi, W, 0)
   = \mathcal{L}_{\mathrm{DPO}}(\pi),
\]
where
$\mathcal{L}_{\mathrm{DPO}}(\pi)
 := -\,\mathbb{E}_{(x,\,y_+,\,y_-)\sim\mathcal{D}}
    \bigl[\log \sigma\bigl(S^{\mathrm{DPO}}_{+-}(x)\bigr)\bigr]$
is the population DPO objective~\eqref{eq:logit}, with $\mathcal{D}$ the
preference-triple distribution of~(5) and $S^{\mathrm{DPO}}_{+-}$
the DPO score of Section~II-B evaluated at the chosen--rejected
pair $(y_+,\,y_-)$.
\end{corollary}

The result follows from the definitions in Section~\ref{sec:method}. When \(\boldsymbol{\gamma}=\mathbf{0}\), the skew-symmetric matrix \(J(\boldsymbol{\gamma})\) in~\eqref{eq:J} is the zero matrix. Hence the curl score \(c_{ij}=\mathbf{v}_i^\top J(\boldsymbol{\gamma})\mathbf{v}_j\) in~\eqref{eq:curl} vanishes for every comparison pair. The penalty term \(\lambda_{\mathrm{reg}}\|\boldsymbol{\gamma}\|_1\) in~\eqref{eq:shpo} also vanishes. Therefore, the SHPO logit in~\eqref{eq:logit} reduces to the DPO logit, and every DPO score can be recovered by choosing \(\boldsymbol{\gamma}=\mathbf{0}\).

We next examine whether this inclusion is strict. The following lemma gives a simple realizability property of the shared projection used in the curl head.

\begin{lemma}\label{lem:realizability}
Let \(\mathbf{h}_A,\mathbf{h}_B,\mathbf{h}_C\in\mathbb{R}^d\) be linearly independent. For any target unit vectors
\[
\mathbf{t}_A,\mathbf{t}_B,\mathbf{t}_C\in\mathbb{S}^{2K-1},
\]
there exists a shared projection \(W\in\mathbb{R}^{d\times 2K}\) such that
\[
W^\top \mathbf{h}_i=\mathbf{t}_i,
\qquad i\in\{A,B,C\}.
\]
After the normalization step in the curl head, the corresponding curl embeddings satisfy \(\mathbf{v}_i=\mathbf{t}_i\) for \(i\in\{A,B,C\}\).

Here $h_A$, $h_B$, and $h_C$ are fixed vectors, they are instantiated as the pooled hidden states in Theorem~\ref{thm:strict-inclusion}.
\end{lemma}

The proof is given in Appendix~\ref{app:proof-realizability}. The lemma is used to ensure that the curl head can realize prescribed directions on a response triangle whenever the corresponding hidden states are nondegenerate.

Fix a prompt $x$ with comparison graph $G=(V,E)$. Let $\Pi$ denote
the set of all policies with full support on $V$. Formally,
\begin{equation*}
\Pi=\{\pi \,:\, \pi(y\mid x)>0 \ \text{for all } y\in V\}.
\end{equation*}
The trainable family in Section~\ref{sec:experiments} is a parametric subset of $\Pi$.
The score classes below are built from policies in $\Pi$. The results
in this section are stated for these classes.
Define
\[
\mathcal{H}_{\mathrm{DPO}}
=
\bigl\{(S^{\pi}_{ij})_{(i,j)\in E}\mid \pi\in\Pi\bigr\},
\]
where \(S^{\pi}_{ij}\) is the DPO edge score defined in Proposition~\ref{prop:dpo-gradient}. Define the SHPO score class by

\begin{equation*}
    \begin{aligned}
       & \mathcal{H}_{\mathrm{SHPO}}
=
\Bigl\{
\bigl(S^{\pi}_{ij}
+
\alpha\,\mathbf{v}_i^\top J(\boldsymbol{\gamma})\mathbf{v}_j
\bigr)_{(i,j)\in E}
:\\
&\pi\in\Pi,\,
W\in\mathbb{R}^{d\times 2K},\,
\boldsymbol{\gamma}\in\mathbb{R}^K
\Bigr\}.
    \end{aligned}
\end{equation*}
Here \(J(\boldsymbol{\gamma})\) is the block-diagonal skew-symmetric matrix in~\eqref{eq:J}, and \(\mathbf{v}_i\) is the normalized projected representation of response \(y_i\). We assume that the relevant log-ratios are finite and that the projected vectors used by the curl head are nonzero.

\begin{theorem}\label{thm:strict-inclusion}
Fix \(K\geq 1\) and \(\alpha\neq 0\). Suppose that the comparison graph contains a triangle \((A,B,C)\). Assume further that the hidden states
\[
\mathbf{h}_A^{\pi_{\mathrm{ref}}},
\mathbf{h}_B^{\pi_{\mathrm{ref}}},
\mathbf{h}_C^{\pi_{\mathrm{ref}}}
\]
under the reference policy are linearly independent, and that all policies under consideration assign positive probability to the responses in \(V\). Then
\[
\mathcal{H}_{\mathrm{DPO}}
\subsetneq
\mathcal{H}_{\mathrm{SHPO}}.
\]
\end{theorem}

The proof is given in Appendix~\ref{app:proof-strict-inclusion}. The inclusion follows from Corollary~\ref{cor:dpo-special}. For strictness, Lemma~\ref{lem:realizability} allows the curl embeddings of \(A,B,C\) to be chosen so that
\[
c_{AB}+c_{BC}+c_{CA}\neq 0.
\]
By Proposition~\ref{prop:dpo-gradient}, every DPO score is a gradient flow. Hence, on the same triangle,
\[
S^{\pi}_{AB}+S^{\pi}_{BC}+S^{\pi}_{CA}=0.
\]
The SHPO score constructed in the proof has nonzero triangular circulation and therefore cannot be represented by any DPO score. Changing the sign of \(\alpha\) reverses the orientation of the constructed cycle, but does not affect the strict-inclusion argument.

The assumptions in Theorem~\ref{thm:strict-inclusion} separate the graph and representation requirements. The triangle \((A,B,C)\) provides a closed three-edge comparison structure on which the quantity \(S_{AB}+S_{BC}+S_{CA}\) is defined. The linear independence of \(\mathbf{h}_A^{\pi_{\mathrm{ref}}}\), \(\mathbf{h}_B^{\pi_{\mathrm{ref}}}\), and \(\mathbf{h}_C^{\pi_{\mathrm{ref}}}\) ensures that the shared projection \(W\) has enough degrees of freedom to realize a nonzero skew-symmetric bilinear score on this triangle. We use the positivity assumption only to keep the DPO log-ratios
finite. Standard softmax language models satisfy this assumption on finite response sets.

We now turn to the \(\ell_1\) penalty on \(\boldsymbol{\gamma}\). In the following result, the policy \(\pi\), the reference policy \(\pi_{\mathrm{ref}}\), and the curl embeddings \(\mathbf{v}\) are fixed. The only parameter to be estimated is then the vector of curl coefficients. For the \(t\)-th comparison, define
\[
\mathbf{X}_t=(X_{t1},\ldots,X_{tK})^\top,
\]
where
\[
X_{t\ell}
=
v^{(2\ell-1)}_{i_t}v^{(2\ell)}_{j_t}
-
v^{(2\ell)}_{i_t}v^{(2\ell-1)}_{j_t},
\qquad \ell=1,\ldots,K.
\]
Let \(h_t\) denote the fixed DPO offset for this comparison. The SHPO logit can be written as
\[
\eta_t(\boldsymbol{\gamma})
=
h_t+\alpha \mathbf{X}_t^\top\boldsymbol{\gamma}.
\]
Thus, with \((\pi,\mathbf{v})\) fixed, estimating \(\boldsymbol{\gamma}\) reduces to an \(\ell_1\)-penalized logistic regression with a fixed offset.

Let \(\mathbf{X}\in\mathbb{R}^{n\times K}\) be the design matrix with rows \(\mathbf{X}_t^\top\). Let
\[
S=\operatorname{supp}(\boldsymbol{\gamma}^*),
\qquad
k^*=|S|,
\]
and define the sparse cone
\[
\mathcal{C}(S,3)
=
\left\{
\boldsymbol{\delta}:
\|\boldsymbol{\delta}_{S^c}\|_1
\leq
3\|\boldsymbol{\delta}_{S}\|_1
\right\}.
\]

\begin{proposition}\label{prop:sparse}
Fix the policy \(\pi\) and the curl embeddings \(\mathbf{v}\), and assume \(\alpha\neq0\). Suppose that, conditional on the realized design \((\mathbf{X}_t,h_t)_{t=1}^n\), the preference labels are independent and follow the logistic SHPO model with true coefficient vector \(\boldsymbol{\gamma}^*\). Assume that \(\boldsymbol{\gamma}^*\) is \(k^*\)-sparse, the bilinear features satisfy \(\max_{t,\ell}|X_{t\ell}|\leq L\), the offsets satisfy \(\max_t |h_t|\leq H\), and the design satisfies the restricted eigenvalue condition
\[
\frac{1}{n}\|\mathbf{X}\boldsymbol{\delta}\|_2^2
\geq
\kappa\|\boldsymbol{\delta}\|_2^2,
\qquad
\boldsymbol{\delta}\in\mathcal{C}(S,3),
\]
for some \(\kappa>0\). Assume further that the logistic curvature is uniformly bounded below on the line segment joining \(\boldsymbol{\gamma}^*\) and \(\widehat{\boldsymbol{\gamma}}\), with lower bound \(w_0>0\). If
\[
\lambda_{\mathrm{reg}}
=
c_\lambda |\alpha|L\sqrt{\frac{\log K}{n}}
\]
for a sufficiently large numerical constant \(c_\lambda\), then, with high probability,
\[
\|\widehat{\boldsymbol{\gamma}}-\boldsymbol{\gamma}^*\|_2
\leq
C
\frac{L}{|\alpha|w_0\kappa}
\sqrt{\frac{k^*\log K}{n}},
\]
where \(C>0\) is a universal constant. Consequently, when \(\alpha\), \(L\), \(w_0\), and \(\kappa\) are treated as fixed constants,
\[
\|\widehat{\boldsymbol{\gamma}}-\boldsymbol{\gamma}^*\|_2
=
O\left(
\sqrt{\frac{k^*\log K}{n}}
\right).
\]
\end{proposition}

The proof is given in Appendix~\ref{app:proof-sparse}. Proposition~\ref{prop:sparse} explains why the $\ell_1$ penalty helps. The result takes the policy and the curl embeddings as fixed. The estimation error matches the standard rate for sparse logistic regression~\cite{buhlmann2011statistics}. Many results in high-dimensional statistics fix the representations in this way. This choice lets us focus on the part of the loss where the penalty acts. A full theory for the joint training of $\pi$, $W$, and $\gamma$ is a natural next step. We leave it as future work. The experiments in Section~\ref{sec:experiments} test the full method in practice.


\section{Experiments}\label{sec:experiments}

This section presents a comprehensive empirical comparison between the proposed SHPO alignment algorithm and other within-BT or Beyond-BT competing approaches,
including preference-accuracy comparisons, ablation studies, and diagnostics of the learned curl head, and generation-quality assessments.

\subsection{Experimental Setup}\label{sec:experimental-setup}

\paragraph{Base Model}
We fine-tune Qwen2.5-7B-Instruct~\cite{qwen2024} using Low-Rank Adaptation
(LoRA~\cite{hu2022lora}; implemented via PEFT~\cite{peft2022}) with rank $r{=}16$ and scaling factor $\alpha_{\textup{LoRA}}{=}32$, under 4-bit
quantization (QLoRA)~\cite{dettmers2023} and bfloat16 precision. The base model has a hidden dimension 
of $d{=}3{,}584$. 
For a fair comparison, the pre-trained Qwen2.5-7B checkpoint is used as  the frozen
reference model $\pi_{\textup{ref}}$
for the proposed SHPO and other competing methods on different datasets. 

\paragraph{Datasets}
Three preference datasets, including HH-RLHF~\cite{bai2022}
($10{,}000$ train / $1{,}000$ test pairs), UltraFeedback~\cite{cui2024ultrafeedback}
(matched $10{,}000$ / $1{,}000$), and a synthetic preference dataset
($10{,}000$ / $1{,}000$), are used to assess the performances of the proposed SHPO and other competing methods. Each example consists of prompts, chosen responses, and rejected responses. 
HH-RLHF and UltraFeedback are available at the links provided in the papers, and we use their pairwise preference versions. 
Due to the low proportion of local cycles in these two datasets, we additionally generate a synthetic dataset from prompt-conditioned four-response sets, with a local cycle injection rate of 0.1 and a label-flip probability of 0.1. The train-test split of this dataset is done at the prompt level rather than the response-pair level.

\begin{figure*}[!t]
    \centering
    \includegraphics[width=0.95\textwidth]{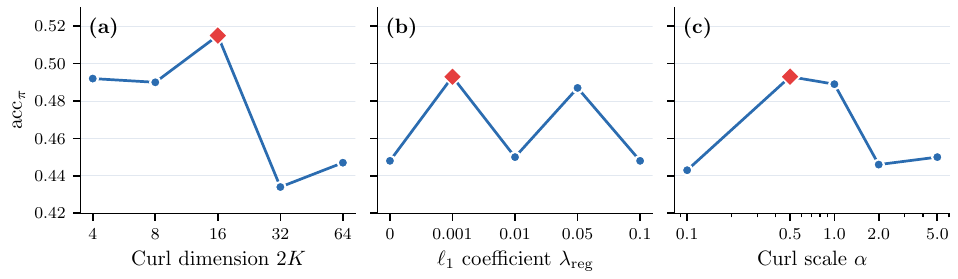}
    \caption{Hyperparameter sensitivity of SHPO on HH-RLHF. 
    (a)~Curl dimension $2K$: best at $16$. 
    (b)~$\ell_1$ coefficient $\lambda_{\textup{reg}}$: best at $0.001$. 
    (c)~Curl scale $\alpha$: best at $0.5$. 
    All results are under single random seed.} 
    \label{fig:sensitivity}
\end{figure*}

\paragraph{Training}
SHPO-specific hyperparameters are $K{=}8$ curl
planes (curl dimension $2K{=}16$), curl scale $\alpha{=}0.5$, and
$\ell_1$ regularization parameter $\lambda_{\textup{reg}}{=}0.001$, which are selected via grid
search (see Figure~\ref{fig:sensitivity} 
and
Appendix~\ref{app:additional-details}) 
All experiments are conducted using Accelerate~\cite{accelerate2022} for distributed training orchestration.
Appendix Table~\ref{tab:experiment-config} provides other key configuration.
The main results and ablations are obtained under $3$ random seeds ($42$, $123$, $456$), and the hyperparameter sweeps and pilot diagnostics use a single seed ($42$). 

\paragraph{Implementation and budget}
All
experiments operate under a constrained training budget ($10{,}000$
preference pairs per dataset, $3$ epochs, QLoRA 4-bit),
to isolate algorithmic contributions under matched resource
constraints. 

\paragraph{Evaluation metrics}

\quad 

\noindent\textbf{Policy accuracy}
($\mathrm{acc}_\pi$) measures the fraction of test pairs where the trained
policy assigns higher log-probability to the chosen response:
$\mathrm{acc}_\pi = \mathbb{E}[\mathbf{1}(\log\pi(y_+\mid x) >
\log\pi(y_-\mid x))]$. 

\noindent\textbf{DPO logit accuracy} ($\mathrm{acc}_{\textup{dpo}}$) measures the
fraction of test pairs where the DPO logit
$\beta(\log\frac{\pi(y_+\mid x)}{\pi_{\textup{ref}}(y_+\mid x)} -
\log\frac{\pi(y_-\mid x)}{\pi_{\textup{ref}}(y_-\mid x)})$ is positive.

\noindent\textbf{Combined accuracy} ($\mathrm{acc}_{\textup{dpo+c}}$) measures the
fraction of test pairs where the full augmented logit $\ell_{+-}$ of
Eq.~\eqref{eq:logit}, including the curl score $\alpha\,c_{+-}$, is
positive. 

The first two metrics are commonly used in literature on preference learning \cite{chen2024preference,lambert2025rewardbench}. For methods without a curl head,
$\mathrm{acc}_{\textup{dpo+c}} = \mathrm{acc}_{\textup{dpo}}$ by
construction.
We use
$\mathrm{acc}_{\pi}$ as the primary metric, and $\mathrm{acc}_{\textup{dpo}}$
and $\mathrm{acc}_{\textup{dpo+c}}$ as the auxiliary  to
separate the transitive and curl contributions.

\paragraph{Competing baselines}
We compare SHPO with five competitors: DPO~\cite{rafailov2023},
IPO~\cite{azar2024} with target margin $\tau{=}2.0$,
cDPO~\cite{mitchell2023} with conservative noise level $0.2$,
rDPO~\cite{chowdhury2024} with label smoothing $\varepsilon{=}0.1$, and
GPO \cite{zhang2025gpm}. 
The non-sparse version of the proposed SHPO is referred to HPO, that is, SHPO with $\lambda_{\textup{reg}}{=}0$, and will be included in the ablation analysis in Section \ref{sec:ablation-studies}. 
Unless otherwise specified, all the above methods share identical base model, LoRA settings, and training hyperparameters. 



\subsection{Main Results}\label{sec:main-results}

Table~\ref{tab:main-results} summarizes preference accuracy results across all six
methods for three datasets. Results are reported as mean~$\pm$~std over three random seeds. Among these six methods, SHPO achieves the highest preference accuracy  for all these datasets in terms of $\mathrm{acc}_\pi$,  $\mathrm{acc}_{\textup{dpo}}$ and $\mathrm{acc}_{\textup{dpo+c}}$.
Since the two real-world datasets, HH-RLHF and UltraFeedback, exhibit more complex and noisy preference structures than the synthetic dataset, all baselines remain modest under the matched low-resource budget listed in Section~\ref{sec:experimental-setup}.
SHPO outperforms the runners-up by $+5.2$ and $+2.5$ percentage points in $\mathrm{acc}_\pi$  on HH-RLHF and UltraFeedback datasets, respectively. 

Compared to DPO, the gain of $\mathrm{acc}_\pi$ on HH-RLHF and UltraFeedback datasets further increases to 5.6 and 9.6 percentage points, respectively.
Particularly for SHPO, its $\mathrm{acc}_{\textup{dpo+c}}$  
exceeds $\mathrm{acc}_{\textup{dpo}}$ by $+1.6$, $+1.0$, and $+2.1$ percentage points
across the three datasets. 
These results suggest that the curl score  carries discriminative signal beyond the DPO logit, and that the incorporation of curl residual can improve preference accuracy. 

Additionally, GPO as a beyond-BT method 
performs at best comparably to the BT family on the
real datasets, but it substantially worse on Synthetic dataset ($\mathrm{acc}_\pi$ $0.543$ vs. $0.784$ for DPO). 
SHPO 
improves $\mathrm{acc}_\pi$ to $0.820 \pm 0.017$ ($+3.6$ percentage over DPO) for the synthetic dataset.
These results suggest that the performance gains arise from the combined use of the curl residual and the $\ell_1$ penalty. 
This finding is further validated by the ablation study in Section~\ref{sec:ablation-studies}.
 

\begin{table*}[!t]
\centering
\caption{Main results on three preference datasets using the proposed SHPO and other competing methods. 
$\mathrm{acc}_\pi$ is policy log-probability
accuracy; $\mathrm{acc}_{\textup{dpo}}$ and $\mathrm{acc}_{\textup{dpo+c}}$ measure the
DPO logit and the combined (DPO $+$ curl) logit respectively. All methods report
3-seed mean $\pm$ std.
\textbf{Bold}: best method per dataset and metric; \underline{underline}: second best.}
\label{tab:main-results}
\footnotesize
\setlength{\tabcolsep}{3.5pt}
\renewcommand{\arraystretch}{1.05}
\begin{adjustbox}{max width=\textwidth,center}
\begin{tabular}{l ccc ccc ccc}
\toprule
\multirow{2}{*}{Method}
 & \multicolumn{3}{c}{HH-RLHF}
 & \multicolumn{3}{c}{UltraFeedback}
 & \multicolumn{3}{c}{Synthetic} \\
\cmidrule(lr){2-4}\cmidrule(lr){5-7}\cmidrule(lr){8-10}
 & $\mathrm{acc}_\pi$ & $\mathrm{acc}_{\textup{dpo}}$ & $\mathrm{acc}_{\textup{dpo+c}}$
 & $\mathrm{acc}_\pi$ & $\mathrm{acc}_{\textup{dpo}}$ & $\mathrm{acc}_{\textup{dpo+c}}$
 & $\mathrm{acc}_\pi$ & $\mathrm{acc}_{\textup{dpo}}$ & $\mathrm{acc}_{\textup{dpo+c}}$ \\
\midrule
DPO   & \sig{0.449}{0.049} & \sig{\underline{0.476}}{0.032} & \sig{\underline{0.476}}{0.032}
      & \sig{0.436}{0.009} & \sig{0.424}{0.114} & \sig{0.424}{0.114}
      & \sig{\underline{0.784}}{0.017} & \sig{0.768}{0.020} & \sig{0.768}{0.020} \\
IPO   & \sig{0.448}{0.061} & \sig{0.443}{0.041} & \sig{0.443}{0.041}
      & \sig{0.503}{0.110} & \sig{0.526}{0.097} & \sig{0.526}{0.097}
      & \sig{0.778}{0.101} & \sig{\underline{0.780}}{0.087} & \sig{\underline{0.780}}{0.087} \\
cDPO  & \sig{0.436}{0.091} & \sig{0.441}{0.036} & \sig{0.441}{0.036}
      & \sig{0.436}{0.047} & \sig{0.438}{0.066} & \sig{0.438}{0.066}
      & \sig{0.740}{0.017} & \sig{0.750}{0.021} & \sig{0.750}{0.021} \\
rDPO  & \sig{0.452}{0.040} & \sig{0.443}{0.066} & \sig{0.443}{0.066}
      & \sig{\underline{0.507}}{0.025} & \sig{\underline{0.579}}{0.051} & \sig{\underline{0.579}}{0.051}
      & \sig{0.772}{0.017} & \sig{0.778}{0.023} & \sig{0.778}{0.023} \\
GPO   & \sig{\underline{0.453}}{0.001} & \sig{0.466}{0.002} & \sig{0.466}{0.002}
      & \sig{0.445}{0.005} & \sig{0.422}{0.004} & \sig{0.422}{0.004}
      & \sig{0.543}{0.004} & \sig{0.541}{0.005} & \sig{0.541}{0.005} \\
\textbf{SHPO}
      & \sig{\textbf{0.505}}{0.007} & \sig{\textbf{0.593}}{0.018} & \sig{\textbf{0.609}}{0.010}
      & \sig{\textbf{0.532}}{0.010} & \sig{\textbf{0.646}}{0.030} & \sig{\textbf{0.656}}{0.014}
      & \sig{\textbf{0.820}}{0.017} & \sig{\textbf{0.800}}{0.027} & \sig{\textbf{0.821}}{0.020} \\
\bottomrule
\end{tabular}
\end{adjustbox}
\end{table*}

\subsection{Ablation Studies}\label{sec:ablation-studies}

\begin{figure*}[!t]
    \centering
    \includegraphics[width=\textwidth]{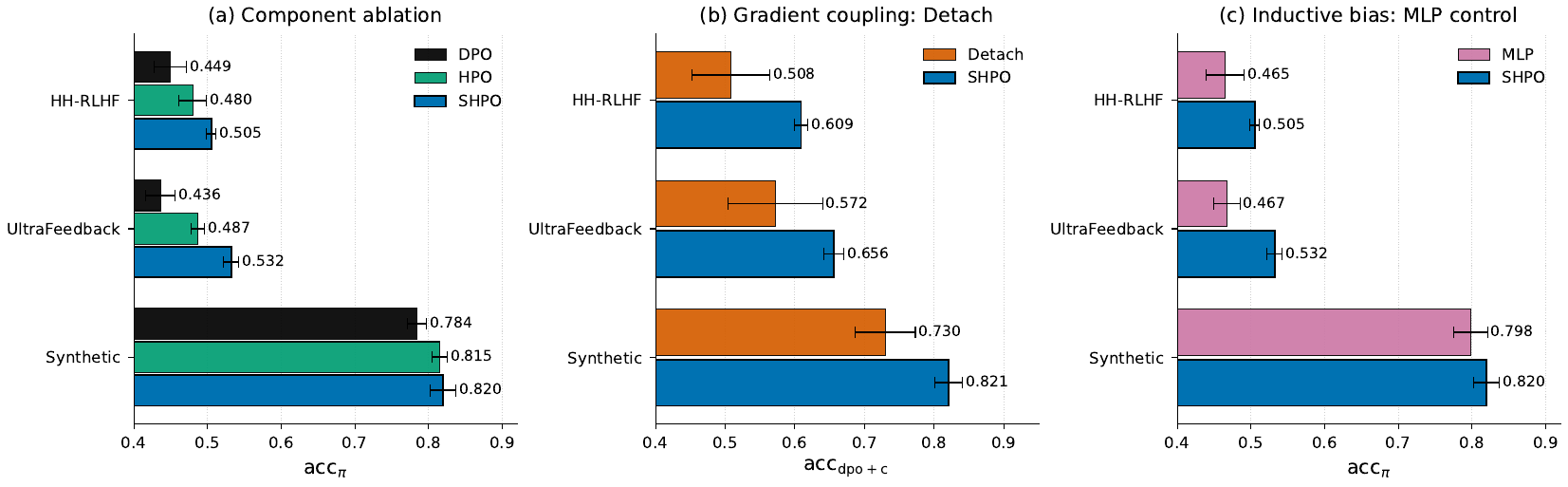}
    \caption{Ablation studies across the three datasets. Bars report $3$-seed means with sample-std error bars.
    (a)~Component ablation DPO vs. HPO  vs. SHPO. 
    (b)~Gradient-coupling ablation 
    SHPO vs. Detach variant. 
    (c)~Inductive-bias control curl head vs. 
    antisymmetrized MLP head. 
    } 
    \label{fig:ablation}
\end{figure*}

We isolate SHPO's contributions through three ablation studies: a) a component
ablation that traces the individual contributions of the curl head and $\ell_1$
regularization, b) a detach ablation that examines the importance of gradient
coupling, and c) a control experiment that replaces the bilinear curl head with a parameter-matched 
MLP head. 
Figure~\ref{fig:ablation} presents these results, while the corresponding numerical values are reported in Table~\ref{tab:ablation}.

\paragraph{Component ablation}
HPO refers to the variation of our SHPO but without $\ell_1$ regularization.
The progression from DPO to HPO and finally to SHPO, shown in 
Figure~\ref{fig:ablation}(a), reveals the individual contributions of the curl head and $\ell_1$ regularization. 
Here, we report $\mathrm{acc}_\pi$ because it reflects end-to-end policy quality. 
The improvement is evident from DPO to HPO and from HPO to SHPO for across all three datasets. The gains are particularly pronounced on the two real-world datasets, whose preference structures are more complex and noisy than those of the synthetic dataset.

\paragraph{Detach ablation}

As discussed in Section \ref{sec:computational details}, in our SHPO method, the shared backbone is updated jointly by gradients from both the gradient and curl components. As an ablation, we block curl-head gradients from updating the shared backbone using a stop-gradient operation.
Figure~\ref{fig:ablation}(b) displays the difference our SHPO and the detach variation on the three datasets.   
Here, we report $\mathrm{acc}_{\textup{dpo+c}}$ because it
captures the joint effect of the DPO logit and curl score without being affected by 
gradient coupling. 
SHPO achieves higher $\mathrm{acc}_{\textup{dpo+c}}$ than its detach variation across these three datasets. It suggests that allowing curl gradients to update the shared backbone is beneficial.

\paragraph{MLP control}
We replace the curl head used in our SHPO 
by an antisymmetrized two-layer MLP head with a matched parameter count. 
As shown in Figure~\ref{fig:ablation}(c), SHPO outperforms this MLP for these three datasets. 
This indicates that the gains attribute to the antisymmetric bilinear structure rather than increased model capacity alone.

\subsection{Curl Head Analysis}

\begin{figure*}[!t]
    \centering
    \includegraphics[width=0.95\textwidth]{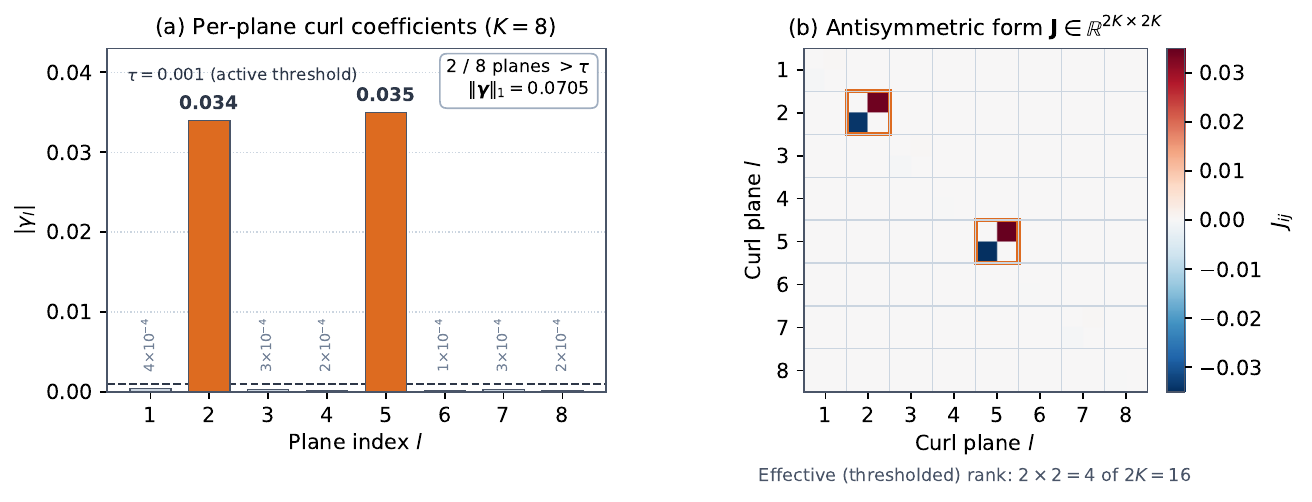}
    \caption{Sparse, low-rank structure exhibited by the $\ell_1$-regularized
    curl head on the pilot checkpoint.}
    \label{fig:sparsity}
\end{figure*}

\paragraph{Sparsity of the learned head} 
For the Synthetic pilot checkpoint with $K=8$ curl planes, only 2 planes
exceed the threshold $\tau=10^{-3}$ (Figure~\ref{fig:sparsity}(a)). 
Of the aggregate norm $\lVert\gamma\rVert_1=0.0705$, 98\% (0.069) lies in the two dominant planes (Table~S.6(a)). 
In view of $J(\gamma)$ given in Eq.\eqref{eq:J}, thresholding leads to its antisymmetric form of rank 4 
(Figure~\ref{fig:sparsity}(b)). 
As stated in Proposition~\ref{prop:sparse}, the estimation loss $O(\sqrt{k^{*}\log K/n})$ rate is favorable in this case wih $k^{*}=2$ and $n=10^4$. 

Table~S.6(b) quantifies the relative contribution of the curl component to the final decision score on the HH-RLHF test set. The mean contribution ratio is 18.13\% and the median is 7.90\%, indicating that the curl component typically acts as a residual correction to the dominant DPO signal. Nevertheless, its contribution becomes substantial for a subset of examples, reaching 36.64\% at the 90th percentile, and exceeding 50\% for 8.73\% of the test samples. This suggests that the curl component is selectively activated for examples exhibiting stronger non-transitive preference patterns. Thus, the curl head is sparse in parameters and selective across pairs.



\paragraph{Recovery of cyclic structure} 
For the Synthetic data with known local cycles, we enumerate all
794 response triangles of the 200-prompt cycle-analysis split in Appendix Table~\ref{tab:curl-diagnostics}(c). 
Among them, 21.28\%
are cyclic ($A\succ B\succ C\succ A$). This fraction exceeds the designed 0.1
injection ratio because the 
0.1 label-flip noise causes additional cyclic triangles. 
The learned
curl head agrees with the observed cycle orientation on 76.4\% of them
(Appendix Table~\ref{tab:curl-diagnostics}(c)). The head thus captures, from the same
$10^4$-pair budget, a signal that is provably outside the reach of
DPO. 

\subsection{Generation Evaluation}\label{sec:generation-eval}

So far, we have demonstrated the preference accuracy gains from the proposed SHPO.  
It is still worth noting that $\mathrm{acc}_\pi$  is computed from unnormalized sequence log-probabilities and is therefore sensitive to response length. As an offline preference metric, it does not directly measure the quality of generated responses. We therefore conduct a generation evaluation using an GPT-4o judge to assess the practical effectiveness of the aligned models in this subsection.

Following the LLM-as-judge protocol \cite{zheng2023judging}, we
generate responses from each trained model on $N=885$ prompts drawn
from AlpacaEval \cite{li2023alpacaeval} and Vicuna-Bench
\cite{chiang2023vicuna}. GPT-4o scores each response on a 1-10 scale according to helpfulness, relevance, accuracy, depth, creativity, and level of detail. The evaluation prompt requests a brief rationale followed by the numerical score. Table~\ref{tab:generation} compares DPO, GPO and SHPO methods in terms of open-ended generation quality. The proposed SHPO method achieves the highest scores across the overall mean, AlpacaEval, and Vicuna-Bench evaluations (Table~\ref{tab:generation}(a)). 

We additionally conduct pairwise generation evaluation on the 805 AlpacaEval prompts.
Each response pair is evaluated twice with reversed presentation orders to reduce position bias. A model is declared the winner only when it is preferred in both orderings; otherwise, the outcome is recorded as a tie.
The proposed SHPO achieves highest win rates against DPO and GPO in the head-to-head comparisons (Table~\ref{tab:generation}(b)-(c)), with both comparisons reaching statistical significance under a two-sided exact binomial test (\emph{p}$<0.001$).

\begin{table*}[!t]
\centering
\caption{Generation evaluation using GPT-4o as the judge. 
Panel~(a): pointwise quality scores 
on
$N{=}885$ prompts from AlpacaEval and Vicuna-Bench, with marginal 
$95\%$ bootstrap confidence intervals (1000 resamples) for each method. 
Panels~(b) and~(c): pairwise head-to-head comparisons.}
\label{tab:generation}
\small
\setlength{\tabcolsep}{5pt}
\begin{tabular}{l c c c}
\toprule
\multicolumn{4}{l}{\textit{(a) Pointwise quality scores (GPT-4o, scale $1$--$10$)}} \\
\midrule
Method         & Overall Mean \; [95\% CI] & AlpacaEval & Vicuna-Bench \\
\midrule
DPO (baseline) & $5.11$ {\scriptsize $[4.94,\, 5.28]$} & $5.03$ & $5.89$ \\
GPO            & $5.89$ {\scriptsize $[5.74,\, 6.04]$} & $5.82$ & $6.59$ \\
\textbf{SHPO}  & $\mathbf{6.31}$ {\scriptsize $[6.12,\, 6.50]$} & $\mathbf{6.24}$ & $\mathbf{7.01}$ \\
\midrule
\multicolumn{4}{l}{\textit{(b) Pairwise preference (SHPO vs.\ DPO, $n{=}805$)}} \\
\midrule
 & SHPO wins & DPO wins & Ties \\
\midrule
Count (\%)
    & $\mathbf{460}\ (\mathbf{57.1\%})$
    & $192\ (23.9\%)$
    & $153\ (19.0\%)$ \\
\multicolumn{4}{r}{\small Win ratio (SHPO\,:\,DPO) $= 2.40{:}1$} \\
\midrule
\multicolumn{4}{l}{\textit{(c) Pairwise preference (SHPO vs.\ GPO, $n{=}805$)}} \\
\midrule
 & SHPO wins & GPO wins & Ties \\
\midrule
Count (\%)
    & $\mathbf{349}\ (\mathbf{43.4\%})$
    & $259\ (32.2\%)$
    & $197\ (24.5\%)$ \\
\multicolumn{4}{r}{\small Win ratio (SHPO\,:\,GPO) $= 1.35{:}1$} \\
\bottomrule
\end{tabular}
\end{table*}

\section{Conclusion and Discussion}


The proposed alignment method SHPO goes beyond transitive preference modeling by preserving dominant transitive signals while capturing cyclic preferences.
We have established three main theoretical results: 1) SHPO reduces to 
DPO 
when the curl coefficients vanish; 2) SHPO strictly enlarges the DPO hypothesis class on comparison graphs containing response triangles; 3) the estimator for curl coefficients admits the usual sparse logistic-regression rate once the policy and response representations are fixed. 
Despite these results, theoretical guarantees under joint neural training, rather than fixed representations, remain open and warrant further investigation.

Empirically, compared to competing baselines, SHPO improves policy accuracy under a matched QLoRA training budget on HH-RLHF, UltraFeedback, and the synthetic cyclic-preference datasets. The ablation studies further demonstrate that the improvement is tied to both the antisymmetric curl structure and the sparsity penalty, rather than merely to adding extra parameters. Ever under this constrained training budget and an automated judge, the present experiments demonstrate the effectiveness of our proposed method across multiple evaluation metrics. 
Exploring the behavior of SHPO under substantially larger training budgets and human evaluation is left for future research. 

Additionally, 
few curl structures occur in the real-data benchmarks, HH-RLHF and UltraFeedback. Collecting cyclic real-data datasets is an important direction for future work.


\section*{Supplementary Materials}
Supplementary materials provide detailed proofs of the theoretical results, per-seed experimental results, additional ablation and sensitivity analyses, curl-head diagnostics, and the full experimental configuration.

\section*{Data Availability Statement}
The experimental datasets 
HH-RLHF~\cite{bai2022} and UltraFeedback~\cite{cui2024ultrafeedback} are publicly available. The Synthetic dataset used in this study is generated as described in Section \ref{sec:experimental-setup}.

\section*{Declaration Statement}
The authors declare no competing interests.


{\appendices
}

 \bibliographystyle{IEEEtran}

\bibliography{reference}

\end{document}


\maketitle


\FloatBarrier
\appendix
\tableofcontents

\section{Proofs}
\label{app:proofs}

Throughout this section, for a linear map \(A\), we write
\(\operatorname{im}(A)\) and \(\ker(A)\) for its image and kernel. If
\(\mathcal U\) is a subspace of an inner-product space, then
\(\mathcal U^\perp\) denotes its orthogonal complement. The symbol
\(\oplus\) denotes an orthogonal direct sum. The adjoint of \(A\) is
denoted by \(A^*\).

\subsection{Uniqueness of the Hodge decomposition}
\label{app:proof-hodge}

\begin{proof}
Let \(G=(V,E)\) be a finite undirected graph. Define the set of triangles in \(G\) by
\[
T(E)
=
\left\{
\{i,j,k\}\subseteq V
\;\middle|\;
\{i,j\},\{j,k\},\{k,i\}\in E
\right\}.
\]

Let \(\Omega^0(G)\) be the space of functions
\[
f\colon V\to\mathbb R.
\]
Let \(\Omega^1(G)\) be the space of antisymmetric edge flows \(S\)
satisfying
\[
S_{ij}=-S_{ji}
\]
for every \(\{i,j\}\in E\). Let \(\Omega^2(G)\) be the space of
alternating triangle flows \(\Phi\) satisfying
\[
\Phi_{ijk}
=
\Phi_{jki}
=
\Phi_{kij}
=
-\Phi_{jik}
=
-\Phi_{ikj}
=
-\Phi_{kji}
\]
for every \(\{i,j,k\}\in T(E)\).

For \(f,g\in\Omega^0(G)\), \(S,U\in\Omega^1(G)\), and
\(\Phi,\Psi\in\Omega^2(G)\), define
\[
\langle f,g\rangle_0
=
\sum_{i\in V}f_i g_i,
\]
\[
\langle S,U\rangle_1
=
\frac12
\sum_{\substack{i,j\in V\\ \{i,j\}\in E}}
S_{ij}U_{ij},
\]
and
\[
\langle \Phi,\Psi\rangle_2
=
\frac16
\sum_{\substack{i,j,k\in V\\ \{i,j,k\}\in T(E)}}
\Phi_{ijk}\Psi_{ijk}.
\]
The factors \(1/2\) and \(1/6\) account for the two orientations of
each edge and the six orderings of each triangle.

Define
\[
d_0\colon\Omega^0(G)\to\Omega^1(G)
\]
by
\[
(d_0f)_{ij}=f_i-f_j,
\]
and define
\[
d_1\colon\Omega^1(G)\to\Omega^2(G)
\]
by
\[
(d_1S)_{ijk}
=
S_{ij}+S_{jk}+S_{ki}.
\]
For every \(f\in\Omega^0(G)\),
\[
\begin{aligned}
(d_1d_0f)_{ijk}
&=
(f_i-f_j)+(f_j-f_k)+(f_k-f_i) \\
&=0.
\end{aligned}
\]
Hence
\[
d_1d_0=0.
\]

Let \(d_0^*\) and \(d_1^*\) be the adjoints of \(d_0\) and \(d_1\)
with respect to the inner products above. We first determine \(d_0^*\).
For \(f\in\Omega^0(G)\) and \(S\in\Omega^1(G)\),
\[
\begin{aligned}
\langle d_0f,S\rangle_1
&=
\frac12
\sum_{\substack{i,j\in V\\ \{i,j\}\in E}}
(f_i-f_j)S_{ij} \\
&=
\sum_{i\in V}
f_i
\sum_{\substack{j\in V\\ \{i,j\}\in E}}
S_{ij}.
\end{aligned}
\]
Therefore,
\[
(d_0^*S)_i
=
\sum_{\substack{j\in V\\ \{i,j\}\in E}}
S_{ij}.
\]

For \(S\in\Omega^1(G)\) and \(\Phi\in\Omega^2(G)\),
\[
\begin{aligned}
\langle d_1S,\Phi\rangle_2
&=
\frac16
\sum_{\substack{i,j,k\in V\\ \{i,j,k\}\in T(E)}}
(S_{ij}+S_{jk}+S_{ki})\Phi_{ijk} \\
&=
\frac12
\sum_{\substack{i,j\in V\\ \{i,j\}\in E}}
S_{ij}
\sum_{\substack{k\in V\\ \{i,j,k\}\in T(E)}}
\Phi_{ijk}.
\end{aligned}
\]
Thus,
\[
(d_1^*\Phi)_{ij}
=
\sum_{\substack{k\in V\\ \{i,j,k\}\in T(E)}}
\Phi_{ijk}.
\]

For a linear map \(A\) between finite-dimensional inner-product
spaces,
\[
\operatorname{im}(A)^\perp=\ker(A^*).
\]
Applying this identity to \(d_0\) and \(d_1^*\) gives
\[
\operatorname{im}(d_0)^\perp
=
\ker(d_0^*),
\]
and
\[
\operatorname{im}(d_1^*)^\perp
=
\ker(d_1).
\]

Moreover, for \(f\in\Omega^0(G)\) and \(\Phi\in\Omega^2(G)\),
\[
\langle d_0f,d_1^*\Phi\rangle_1
=
\langle d_1d_0f,\Phi\rangle_2
=
0.
\]
Hence
\[
\operatorname{im}(d_0)
\perp
\operatorname{im}(d_1^*).
\]

Define the first Hodge Laplacian by
\[
\Delta_1
=
d_0d_0^*+d_1^*d_1.
\]
For every \(S\in\Omega^1(G)\),
\[
\begin{aligned}
\langle S,\Delta_1S\rangle_1
&=
\langle S,d_0d_0^*S\rangle_1
+
\langle S,d_1^*d_1S\rangle_1 \\
&=
\|d_0^*S\|_0^2+\|d_1S\|_2^2.
\end{aligned}
\]
It follows that
\[
\ker(\Delta_1)
=
\ker(d_0^*)\cap\ker(d_1).
\]

By the finite-dimensional Hodge decomposition~\cite{jiang2011,lim2020},
\[
\Omega^1(G)
=
\operatorname{im}(d_0)
\oplus
\ker(\Delta_1)
\oplus
\operatorname{im}(d_1^*).
\]
Therefore, every \(S\in\Omega^1(G)\) has a unique representation
\[
S
=
S^{\mathrm{grad}}
+
S^{\mathrm{harm}}
+
S^{\mathrm{curl}},
\]
where
\[
S^{\mathrm{grad}}\in\operatorname{im}(d_0),
\qquad
S^{\mathrm{harm}}\in\ker(\Delta_1),
\qquad
S^{\mathrm{curl}}\in\operatorname{im}(d_1^*).
\]

If \(G\) is connected and
\[
S^{\mathrm{grad}}=d_0g,
\]
then \(g\) is unique up to an additive constant, because
\[
\ker(d_0)
=
\{g\in\Omega^0(G)\mid g_i=c \text{ for all }i\in V\}.
\]
The condition
\[
\sum_{i\in V}g_i=0
\]
determines this constant uniquely.
\end{proof}

\subsection{Proof of DPO as a Gradient-Flow Model}
\label{app:proof-dpo-gradient}

\begin{proposition}\label{prop:dpo-gradient}
Fix a prompt $x$ with response set $V$. For any policies $\pi$ and $\pi_{\textup{ref}}$ with full
support on $V$ (suggesting the finiteness of the log-ratios), define
\[
  S_{ij}^{\pi}
  = \beta\!\left(
      \log\frac{\pi(y_i \mid x)}{\pi_{\textup{ref}}(y_i \mid x)}
    - \log\frac{\pi(y_j \mid x)}{\pi_{\textup{ref}}(y_j \mid x)}
  \right).
\]
Then there exists a vertex potential $g\colon V \to \mathbb{R}$ such that
\[
  S_{ij}^{\pi} = g_i - g_j \qquad \text{for all } i,j.
\]
Consequently, every triangle $(i,j,k)$ in the comparison graph satisfies
\[
  S_{ij}^{\pi} + S_{jk}^{\pi} + S_{ki}^{\pi} = 0.
\]
\end{proposition}

\begin{proof}
Fix a prompt \(x\). For each \(i\in V\), define
\[
g_i
=
\beta
\log
\frac{\pi(y_i\mid x)}
{\pi_{\mathrm{ref}}(y_i\mid x)}.
\]
The full-support assumption ensures that \(g_i\) is finite.

By the definition of the DPO edge score~\cite{rafailov2023},
\[
\begin{aligned}
S_{ij}^{\pi}
&=
\beta
\log
\frac{\pi(y_i\mid x)}
{\pi_{\mathrm{ref}}(y_i\mid x)}
-
\beta
\log
\frac{\pi(y_j\mid x)}
{\pi_{\mathrm{ref}}(y_j\mid x)} \\
&=
g_i-g_j.
\end{aligned}
\]
Therefore,
\[
S^\pi=d_0g,
\]
with $d_0$ defined in Section \ref{app:proof-hodge},
and \(S^\pi\) is a gradient flow.

Consider a closed path
\[
i_1\to i_2\to\cdots\to i_m\to i_1
\]
in \(G\). Then
\[
\begin{aligned}
&S_{i_1i_2}^{\pi}
+
S_{i_2i_3}^{\pi}
+\cdots+
S_{i_{m-1}i_m}^{\pi}
+
S_{i_mi_1}^{\pi}
\\
&\quad=
(g_{i_1}-g_{i_2})
+
(g_{i_2}-g_{i_3})
+\cdots+
(g_{i_{m-1}}-g_{i_m})
+
(g_{i_m}-g_{i_1})
\\
&\quad=0.
\end{aligned}
\]
In particular, for every triangle \((i,j,k)\),
\[
S_{ij}^{\pi}+S_{jk}^{\pi}+S_{ki}^{\pi}=0.
\]
\end{proof}

\subsection{Proof of Proposition~\ref{prop:canonical-skew}}
\label{app:proof-canonical-skew}

\begin{proof}
Let
\[
A\in\mathbb R^{2K\times2K}
\]
satisfy
\[
A^\top=-A.
\]
By the real canonical form of skew-symmetric~\cite{hornjohnson2013}, there exists an
orthogonal matrix \(Q\in\mathbb R^{2K\times2K}\) such that
\[
QAQ^\top=J,
\]
where
\[
J
=
\operatorname{diag}
\left(
\begin{pmatrix}
0&\lambda_1\\
-\lambda_1&0
\end{pmatrix},
\ldots,
\begin{pmatrix}
0&\lambda_r\\
-\lambda_r&0
\end{pmatrix},
0_{2(K-r)}
\right)
\]
for some \(r\le K\) and
\[
\lambda_1,\ldots,\lambda_r>0.
\]
Equivalently,
\[
A=Q^\top JQ.
\]

For any
\(\mathbf v_i,\mathbf v_j\in\mathbb R^{2K}\),
\[
\begin{aligned}
\mathbf v_i^\top A\mathbf v_j
&=
\mathbf v_i^\top Q^\top JQ\mathbf v_j \\
&=
(Q\mathbf v_i)^\top J(Q\mathbf v_j).
\end{aligned}
\]
Define
\[
\widetilde{\mathbf v}_i=Q\mathbf v_i.
\]
Then
\[
\mathbf v_i^\top A\mathbf v_j
=
\widetilde{\mathbf v}_i^\top
J
\widetilde{\mathbf v}_j.
\]

Suppose that the embeddings are formed by
\[
\mathbf v_i
=
\frac{W^\top\mathbf h_i}
{\|W^\top\mathbf h_i\|_2}
\]
whenever \(W^\top\mathbf h_i\neq0\). Define
\[
\widetilde W=WQ^\top.
\]
Then
\[
\widetilde W^\top\mathbf h_i
=
QW^\top\mathbf h_i.
\]
Since \(Q\) is orthogonal,
\[
\|\widetilde W^\top\mathbf h_i\|_2
=
\|W^\top\mathbf h_i\|_2.
\]
Hence
\[
\frac{\widetilde W^\top\mathbf h_i}
{\|\widetilde W^\top\mathbf h_i\|_2}
=
Q
\frac{W^\top\mathbf h_i}
{\|W^\top\mathbf h_i\|_2}
=
Q\mathbf v_i
=
\widetilde{\mathbf v}_i.
\]

Thus, when \(W\) is freely trainable, every skew-symmetric bilinear
score can be represented by the block-diagonal form in
Eq.~\eqref{eq:J}.

The canonical form above uses nonnegative coefficients
\(\lambda_\ell\). In the SHPO parameterization, the coefficients
\(\gamma_\ell\) are allowed to take values in \(\mathbb R\). For the
\(\ell\)-th block, the corresponding contribution is
\[
\gamma_\ell
\left(
v_i^{(2\ell-1)}v_j^{(2\ell)}
-
v_i^{(2\ell)}v_j^{(2\ell-1)}
\right).
\]
Changing \(\gamma_\ell\) to \(-\gamma_\ell\) reverses the sign of this
contribution for every pair \((i,j)\). Therefore,
\(\lvert\gamma_\ell\rvert\) controls the magnitude of the contribution,
while the sign of \(\gamma_\ell\) determines its orientation.
\end{proof}

\subsection{Proof of antisymmetry}
\label{app:proof-antisym}

\begin{proof}
For every pair \(i,j\),
\[
c_{ji}
=
\mathbf v_j^\top J\mathbf v_i.
\]
Since \(c_{ji}\) is a scalar,
\[
\begin{aligned}
c_{ji}
&=
\bigl(\mathbf v_j^\top J\mathbf v_i\bigr)^\top \\
&=
\mathbf v_i^\top J^\top\mathbf v_j.
\end{aligned}
\]
Using \(J^\top=-J\),
\[
c_{ji}
=
-\mathbf v_i^\top J\mathbf v_j
=
-c_{ij}.
\]
Thus \(c_{ij}\) is antisymmetric. In particular,
\[
c_{ii}
=
-c_{ii},
\]
and hence
\[
c_{ii}=0.
\]
\end{proof}

\subsection{Proof of Corollary~\ref{cor:dpo-special}}
\label{app:proof-dpo-special}

\begin{proof}
Let
\[
\boldsymbol{\gamma}=\mathbf0.
\]
Then
\[
J(\boldsymbol{\gamma})=0,
\]
and therefore
\[
c_{ij}
=
\mathbf v_i^\top J(\boldsymbol{\gamma})\mathbf v_j
=
0
\]
for every pair \((i,j)\).

The SHPO logit becomes
\[
\ell_{ij}
=
\beta
\left[
\log
\frac{\pi(y_i\mid x)}
{\pi_{\mathrm{ref}}(y_i\mid x)}
-
\log
\frac{\pi(y_j\mid x)}
{\pi_{\mathrm{ref}}(y_j\mid x)}
\right],
\]
which is the DPO logit. Moreover,
\[
\lambda_{\mathrm{reg}}
\|\boldsymbol{\gamma}\|_1
=
0.
\]
Hence
\[
\mathcal L_{\mathrm{SHPO}}(\pi,W,\mathbf0)
=
\mathcal L_{\mathrm{DPO}}(\pi).
\]
\end{proof}

\subsection{Proof of Lemma~\ref{lem:realizability}}
\label{app:proof-realizability}

\begin{proof}
Let
\[
H
=
\begin{pmatrix}
\mathbf h_A&\mathbf h_B&\mathbf h_C
\end{pmatrix}
\in\mathbb R^{d\times3}
\]
and
\[
T
=
\begin{pmatrix}
\mathbf t_A&\mathbf t_B&\mathbf t_C
\end{pmatrix}
\in\mathbb R^{2K\times3}.
\]
Since
\(\mathbf h_A,\mathbf h_B,\mathbf h_C\)
are linearly independent, \(H\) has full column rank. Hence
\[
H^\top H
\]
is invertible.

Define
\[
W
=
H(H^\top H)^{-1}T^\top.
\]
Then
\[
W^\top H
=
T(H^\top H)^{-1}H^\top H
=
T.
\]
Therefore,
\[
W^\top\mathbf h_i
=
\mathbf t_i,
\qquad
i\in\{A,B,C\}.
\]
Since each \(\mathbf t_i\) is a unit vector,
\[
\frac{W^\top\mathbf h_i}
{\|W^\top\mathbf h_i\|_2}
=
\mathbf t_i.
\]
Thus
\[
\mathbf v_i=\mathbf t_i,
\qquad
i\in\{A,B,C\}.
\]
\end{proof}

\subsection{Proof of Theorem~\ref{thm:strict-inclusion}}
\label{app:proof-strict-inclusion}

\begin{proof}
We first prove
\[
\mathcal H_{\mathrm{DPO}}
\subseteq
\mathcal H_{\mathrm{SHPO}}.
\]
Let
\[
(S_{ij}^{\pi})_{(i,j)\in E}
\in
\mathcal H_{\mathrm{DPO}}.
\]
Choose the same policy \(\pi\), an arbitrary projection \(W\), and
\[
\boldsymbol{\gamma}=\mathbf0.
\]
By Corollary~\ref{cor:dpo-special}, the curl term vanishes. Hence
\[
(S_{ij}^{\pi})_{(i,j)\in E}
\in
\mathcal H_{\mathrm{SHPO}}.
\]

It remains to prove that the inclusion is strict. Set
\[
\pi=\pi_{\mathrm{ref}}.
\]
For the remainder of the proof, write
\[
\mathbf h_i
=
\mathbf h_i^{\pi_{\mathrm{ref}}}.
\]
By assumption,
\[
\mathbf h_A,\mathbf h_B,\mathbf h_C
\]
are linearly independent.

Choose
\[
\mathbf v_A
=
(1,0,0,\ldots,0)^\top,
\]
\[
\mathbf v_B
=
\left(
-\frac12,
\frac{\sqrt3}{2},
0,\ldots,0
\right)^\top,
\]
and
\[
\mathbf v_C
=
\left(
-\frac12,
-\frac{\sqrt3}{2},
0,\ldots,0
\right)^\top.
\]
These vectors have unit norm. By Lemma~\ref{lem:realizability}, there exists a projection \(W\) that realizes them as the normalized curl embeddings of \(A,B,C\).

Choose
\[
\gamma_1=1,
\qquad
\gamma_\ell=0,
\quad
\ell=2,\ldots,K.
\]
Then
\[
c_{ij}
=
v_i^{(1)}v_j^{(2)}
-
v_i^{(2)}v_j^{(1)}
\]
on the active curl plane. Direct calculation gives
\[
c_{AB}
=
\frac{\sqrt3}{2},
\qquad
c_{BC}
=
\frac{\sqrt3}{2},
\qquad
c_{CA}
=
\frac{\sqrt3}{2}.
\]
Therefore,
\[
c_{AB}+c_{BC}+c_{CA}
=
\frac{3\sqrt3}{2}.
\]

Since
\[
\pi=\pi_{\mathrm{ref}},
\]
the DPO part of the SHPO score is zero. Hence
\[
S_{ij}^{\mathrm{SHPO}}
=
\alpha c_{ij}.
\]
It follows that
\[
S_{AB}^{\mathrm{SHPO}}
+
S_{BC}^{\mathrm{SHPO}}
+
S_{CA}^{\mathrm{SHPO}}
=
\frac{3\sqrt3}{2}\alpha
\neq0.
\]

By Proposition~\ref{prop:dpo-gradient}, every DPO score has zero circulation on every triangle. Therefore,
\[
(S_{ij}^{\mathrm{SHPO}})_{(i,j)\in E}
\notin
\mathcal H_{\mathrm{DPO}}.
\]
Consequently,
\[
\mathcal H_{\mathrm{DPO}}
\subsetneq
\mathcal H_{\mathrm{SHPO}}.
\]
\end{proof}

\subsection{Proof of Proposition~\ref{prop:sparse}}
\label{app:proof-sparse}

\begin{proof}
Condition on the fixed design
\[
\mathfrak D
=
\{(\mathbf X_t,h_t)\}_{t=1}^n.
\]
For \(t=1,\ldots,n\), let
\[
z_t\in\{-1,1\}
\]
be the preference label and define
\[
\eta_t(\boldsymbol{\gamma})
=
h_t+\alpha\mathbf X_t^\top\boldsymbol{\gamma}.
\]
By assumption,
\[
\mathbb P
\left(
z_t=1
\mid
\mathfrak D
\right)
=
\sigma
\left(
\eta_t(\boldsymbol{\gamma}^*)
\right),
\]
and the labels are conditionally independent.

Define the empirical logistic loss
\[
\widehat{\mathcal L}_n(\boldsymbol{\gamma})
=
\frac1n
\sum_{t=1}^n
\log
\left[
1+
\exp
\left\{
-z_t\eta_t(\boldsymbol{\gamma})
\right\}
\right].
\]
Let
\[
\widehat{\boldsymbol{\gamma}}
\in
\arg\min_{\boldsymbol{\gamma}\in\mathbb R^K}
\left\{
\widehat{\mathcal L}_n(\boldsymbol{\gamma})
+
\lambda_{\mathrm{reg}}
\|\boldsymbol{\gamma}\|_1
\right\}.
\]
Set
\[
\boldsymbol{\Delta}
=
\widehat{\boldsymbol{\gamma}}
-
\boldsymbol{\gamma}^*.
\]

Let
\[
S
=
\operatorname{supp}
(\boldsymbol{\gamma}^*),
\qquad
|S|=k^*.
\]

We first bound the score at
\(\boldsymbol{\gamma}^*\).
Writing
\[
y_t=\frac{1+z_t}{2},
\]
we have
\[
\nabla
\widehat{\mathcal L}_n
(\boldsymbol{\gamma})
=
\frac{\alpha}{n}
\sum_{t=1}^n
\mathbf X_t
\left[
\sigma
\left(
\eta_t(\boldsymbol{\gamma})
\right)
-y_t
\right].
\]
Conditional on \(\mathfrak D\),
\[
\mathbb E
\left[
y_t
\mid
\mathfrak D
\right]
=
\sigma
\left(
\eta_t(\boldsymbol{\gamma}^*)
\right).
\]
Hence each coordinate of
\[
\nabla
\widehat{\mathcal L}_n
(\boldsymbol{\gamma}^*)
\]
is a sum of conditionally independent mean-zero variables.

Since
\[
|X_{t\ell}|\le L,
\]
Hoeffding's inequality~\cite{hoeffding1963} gives
\[
\mathbb P
\left(
\left|
\left[
\nabla
\widehat{\mathcal L}_n
(\boldsymbol{\gamma}^*)
\right]_\ell
\right|
\ge u
\ \middle|\
\mathfrak D
\right)
\le
2
\exp
\left(
-\frac{nu^2}
{2\alpha^2L^2}
\right).
\]
Taking
\[
u
=
2|\alpha|L
\sqrt{\frac{\log K}{n}}
\]
and applying the union bound yields
\[
\left\|
\nabla
\widehat{\mathcal L}_n
(\boldsymbol{\gamma}^*)
\right\|_\infty
\le
2|\alpha|L
\sqrt{\frac{\log K}{n}}
\]
with probability at least \(1-2/K\). Therefore, for
\[
\lambda_{\mathrm{reg}}
=
4|\alpha|L
\sqrt{\frac{\log K}{n}},
\]
we have
\[
\lambda_{\mathrm{reg}}
\ge
2
\left\|
\nabla
\widehat{\mathcal L}_n
(\boldsymbol{\gamma}^*)
\right\|_\infty
\]
on the same event.

Optimality of
\(\widehat{\boldsymbol{\gamma}}\)
gives
\[
\widehat{\mathcal L}_n
(\boldsymbol{\gamma}^*+\boldsymbol{\Delta})
-
\widehat{\mathcal L}_n
(\boldsymbol{\gamma}^*)
\le
-\lambda_{\mathrm{reg}}
\left(
\|\boldsymbol{\gamma}^*
+\boldsymbol{\Delta}\|_1
-
\|\boldsymbol{\gamma}^*\|_1
\right).
\]
By convexity,
\[
\widehat{\mathcal L}_n
(\boldsymbol{\gamma}^*+\boldsymbol{\Delta})
-
\widehat{\mathcal L}_n
(\boldsymbol{\gamma}^*)
\ge
\left\langle
\nabla
\widehat{\mathcal L}_n
(\boldsymbol{\gamma}^*),
\boldsymbol{\Delta}
\right\rangle.
\]
Moreover,
\[
\|\boldsymbol{\gamma}^*
+\boldsymbol{\Delta}\|_1
-
\|\boldsymbol{\gamma}^*\|_1
\ge
\|\boldsymbol{\Delta}_{S^c}\|_1
-
\|\boldsymbol{\Delta}_S\|_1.
\]
It follows that
\[
\left\langle
\nabla
\widehat{\mathcal L}_n
(\boldsymbol{\gamma}^*),
\boldsymbol{\Delta}
\right\rangle
\le
-\lambda_{\mathrm{reg}}
\left(
\|\boldsymbol{\Delta}_{S^c}\|_1
-
\|\boldsymbol{\Delta}_S\|_1
\right).
\]
By Hölder's inequality and the score bound,
\[
\left|
\left\langle
\nabla
\widehat{\mathcal L}_n
(\boldsymbol{\gamma}^*),
\boldsymbol{\Delta}
\right\rangle
\right|
\le
\frac{\lambda_{\mathrm{reg}}}{2}
\|\boldsymbol{\Delta}\|_1.
\]
Combining the preceding two inequalities gives
\[
\|\boldsymbol{\Delta}_{S^c}\|_1
\le
3
\|\boldsymbol{\Delta}_S\|_1.
\]
Thus
\[
\boldsymbol{\Delta}
\in
\mathcal C(S,3).
\]

We next establish restricted strong~\cite{negahban2012unified}. For
\[
\boldsymbol{\delta}
\in
\mathcal C(S,3),
\qquad
\|\boldsymbol{\delta}\|_2\le R,
\]
we have
\[
\|\boldsymbol{\delta}\|_1
\le
4\|\boldsymbol{\delta}_S\|_1
\le
4\sqrt{k^*}
\|\boldsymbol{\delta}\|_2
\le
4\sqrt{k^*}R.
\]
For \(s\in[0,1]\), define
\[
\eta_t^s
=
h_t
+
\alpha
\mathbf X_t^\top
\left(
\boldsymbol{\gamma}^*
+
s\boldsymbol{\delta}
\right).
\]
Using
\[
|h_t|\le H
\]
and
\[
\|\mathbf X_t\|_\infty\le L,
\]
we obtain
\[
|\eta_t^s|
\le
H
+
|\alpha|L
\left(
\|\boldsymbol{\gamma}^*\|_1
+
4\sqrt{k^*}R
\right)
=
B.
\]
Set
\[
w_B
=
\sigma(B)
\{1-\sigma(B)\}.
\]
Then
\[
\sigma(\eta_t^s)
\{1-\sigma(\eta_t^s)\}
\ge
w_B.
\]

Taylor's formula gives
\[
\begin{aligned}
&
\widehat{\mathcal L}_n
(\boldsymbol{\gamma}^*
+\boldsymbol{\delta})
-
\widehat{\mathcal L}_n
(\boldsymbol{\gamma}^*)
-
\left\langle
\nabla
\widehat{\mathcal L}_n
(\boldsymbol{\gamma}^*),
\boldsymbol{\delta}
\right\rangle
\\
&\qquad
=
\frac{\alpha^2}{n}
\int_0^1
(1-s)
\sum_{t=1}^n
w_t(s)
\left(
\mathbf X_t^\top
\boldsymbol{\delta}
\right)^2
\,ds,
\end{aligned}
\]
where
\[
w_t(s)
=
\sigma(\eta_t^s)
\{1-\sigma(\eta_t^s)\}.
\]
Therefore,
\[
\begin{aligned}
&
\widehat{\mathcal L}_n
(\boldsymbol{\gamma}^*
+\boldsymbol{\delta})
-
\widehat{\mathcal L}_n
(\boldsymbol{\gamma}^*)
-
\left\langle
\nabla
\widehat{\mathcal L}_n
(\boldsymbol{\gamma}^*),
\boldsymbol{\delta}
\right\rangle
\\
&\qquad
\ge
\frac{\alpha^2w_B}{2n}
\|\mathbf X\boldsymbol{\delta}\|_2^2.
\end{aligned}
\]
By the restricted eigenvalue condition~\cite{bickel2009simultaneous,wainwright2019high},
\[
\frac1n
\|\mathbf X\boldsymbol{\delta}\|_2^2
\ge
\kappa
\|\boldsymbol{\delta}\|_2^2.
\]
Hence
\[
\widehat{\mathcal L}_n
(\boldsymbol{\gamma}^*
+\boldsymbol{\delta})
-
\widehat{\mathcal L}_n
(\boldsymbol{\gamma}^*)
-
\left\langle
\nabla
\widehat{\mathcal L}_n
(\boldsymbol{\gamma}^*),
\boldsymbol{\delta}
\right\rangle
\ge
\frac{\kappa'}2
\|\boldsymbol{\delta}\|_2^2,
\]
where
\[
\kappa'
=
\alpha^2w_B\kappa.
\]

Assume
\[
\frac{3\sqrt{k^*}
\lambda_{\mathrm{reg}}}
{\kappa'}
<
R.
\]
We claim that
\[
\|\boldsymbol{\Delta}\|_2<R.
\]
Suppose instead that
\[
\|\boldsymbol{\Delta}\|_2\ge R.
\]
Set
\[
\widetilde{\boldsymbol{\Delta}}
=
\frac{R}
{\|\boldsymbol{\Delta}\|_2}
\boldsymbol{\Delta}.
\]
Then
\[
\widetilde{\boldsymbol{\Delta}}
\in
\mathcal C(S,3),
\qquad
\|\widetilde{\boldsymbol{\Delta}}\|_2=R.
\]
Convexity of the penalized objective implies that
\(\widetilde{\boldsymbol{\Delta}}\)
satisfies the basic~\cite{buhlmann2011statistics}. Applying the restricted strong convexity bound gives
\[
\frac{\kappa'}2R^2
\le
\frac{3\lambda_{\mathrm{reg}}}{2}
\sqrt{k^*}R.
\]
Therefore,
\[
R
\le
\frac{3\sqrt{k^*}
\lambda_{\mathrm{reg}}}
{\kappa'},
\]
which contradicts the localization condition. Hence
\[
\|\boldsymbol{\Delta}\|_2<R.
\]

The restricted strong convexity bound may now be applied to
\(\boldsymbol{\Delta}\). Combining it with the basic inequality yields
\[
\frac{\kappa'}2
\|\boldsymbol{\Delta}\|_2^2
\le
\frac{3\lambda_{\mathrm{reg}}}{2}
\|\boldsymbol{\Delta}_S\|_1.
\]
Since
\[
\|\boldsymbol{\Delta}_S\|_1
\le
\sqrt{k^*}
\|\boldsymbol{\Delta}\|_2,
\]
we obtain
\[
\|\boldsymbol{\Delta}\|_2
\le
\frac{3\sqrt{k^*}
\lambda_{\mathrm{reg}}}
{\kappa'}.
\]
Substituting
\[
\lambda_{\mathrm{reg}}
=
4|\alpha|L
\sqrt{\frac{\log K}{n}}
\]
and
\[
\kappa'
=
\alpha^2w_B\kappa
\]
gives
\[
\left\|
\widehat{\boldsymbol{\gamma}}
-
\boldsymbol{\gamma}^*
\right\|_2
\le
\frac{12L}
{|\alpha|w_B\kappa}
\sqrt{
\frac{k^*\log K}{n}
}.
\]
The bound holds with conditional probability at least
\[
1-\frac{2}{K}.
\]
\end{proof}

\FloatBarrier

\section{Per-Seed Results}\label{app:per-seed}

\begin{figure}[!t]
    \centering
    \includegraphics[width=\textwidth]{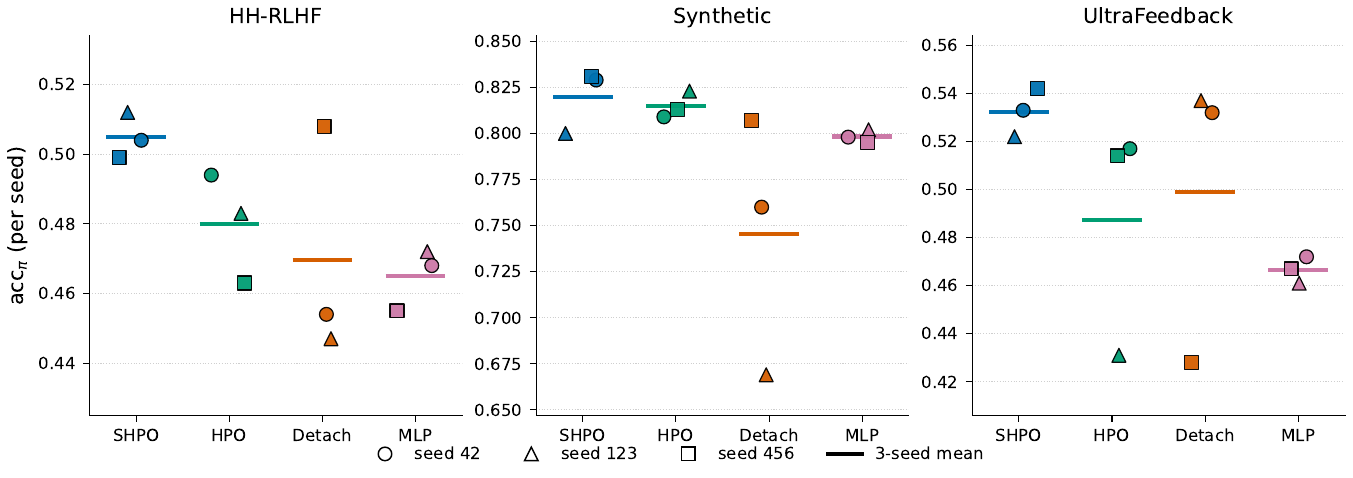}
    \caption{Per-seed pairwise policy accuracy ($\mathrm{acc}_\pi$) for the
    four curl/bilinear-head methods on the three datasets. Each marker is a
    single seed (\(\circ\)~seed~$42$, \(\triangle\)~seed~$123$,
    \(\square\)~seed~$456$); the horizontal bar is the per-method 3-seed
    mean. Tight clusters indicate cross-seed stability (SHPO, MLP); wide
    spreads indicate brittleness (Detach on every dataset, HPO on
    UltraFeedback). Numerical values are listed in
    Table~\ref{tab:per-seed-hh}.}
    \label{fig:per-seed}
\end{figure}

Table~\ref{tab:per-seed-hh} reports individual-seed
results for the four methods with curl or bilinear heads (SHPO, HPO, Detach,
MLP) across all three datasets (HH-RLHF~\cite{bai2022}, Synthetic, and UltraFeedback~\cite{cui2024ultrafeedback}), and Figure~\ref{fig:per-seed} visualizes the
seed-to-seed dispersion. These complement the aggregated summaries in
the main text: mean~$\pm$~std entries in Table~\ref{tab:main-results} and
3-seed mean summaries in Figure~\ref{fig:ablation}.
Per-seed listings are included in this paper only for these four methods;
the DPO~\cite{rafailov2023}, IPO~\cite{azar2024}, cDPO~\cite{mitchell2023},
and rDPO~\cite{chowdhury2024} rows of Table~\ref{tab:main-results} are
aggregated from the same experiment log, whose per-seed listings for these
baselines will be included with the released code and logs per the Data
Availability statement.

\begin{table}[!t]
\centering
\caption{Per-seed results across the three datasets ($10{,}000$ training / $1{,}000$ test pairs).
Each row is a single-seed evaluation.
\textbf{Bold}: best per (dataset, metric) across all methods and seeds.}
\label{tab:per-seed-hh}
\footnotesize
\setlength{\tabcolsep}{3.5pt}
\renewcommand{\arraystretch}{1.05}
\begin{adjustbox}{max width=\textwidth,center}
\begin{tabular}{l c ccc ccc ccc}
\toprule
\multirow{2}{*}{Method} & \multirow{2}{*}{Seed}
 & \multicolumn{3}{c}{HH-RLHF}
 & \multicolumn{3}{c}{Synthetic}
 & \multicolumn{3}{c}{UltraFeedback} \\
\cmidrule(lr){3-5}\cmidrule(lr){6-8}\cmidrule(lr){9-11}
 &  & $\mathrm{acc}_\pi$ & $\mathrm{acc}_{\textup{dpo}}$ & $\mathrm{acc}_{\textup{dpo+c}}$
    & $\mathrm{acc}_\pi$ & $\mathrm{acc}_{\textup{dpo}}$ & $\mathrm{acc}_{\textup{dpo+c}}$
    & $\mathrm{acc}_\pi$ & $\mathrm{acc}_{\textup{dpo}}$ & $\mathrm{acc}_{\textup{dpo+c}}$ \\
\midrule
\multirow{3}{*}{SHPO}
  & 42  & 0.504 & \textbf{0.611} & \textbf{0.619}
        & 0.829 & \textbf{0.830} & 0.831
        & 0.533 & \textbf{0.679} & 0.654 \\
  & 123 & \textbf{0.512} & 0.575 & 0.611
        & 0.800 & 0.793 & 0.799
        & 0.522 & 0.638 & 0.643 \\
  & 456 & 0.499 & 0.595 & 0.599
        & \textbf{0.831} & 0.778 & \textbf{0.835}
        & \textbf{0.542} & 0.621 & \textbf{0.671} \\
\midrule
\multirow{3}{*}{HPO}
  & 42  & 0.494 & 0.606 & 0.611
        & 0.809 & 0.815 & 0.818
        & 0.517 & 0.614 & 0.628 \\
  & 123 & 0.483 & 0.558 & 0.564
        & 0.823 & 0.811 & 0.819
        & 0.431 & 0.409 & 0.410 \\
  & 456 & 0.463 & 0.466 & 0.472
        & 0.813 & 0.794 & 0.797
        & 0.514 & 0.581 & 0.594 \\
\midrule
\multirow{3}{*}{Detach}
  & 42  & 0.454 & 0.471 & 0.477
        & 0.760 & 0.691 & 0.702
        & 0.532 & 0.649 & 0.659 \\
  & 123 & 0.447 & 0.434 & 0.435
        & 0.669 & 0.672 & 0.679
        & 0.537 & 0.643 & 0.650 \\
  & 456 & 0.508 & 0.606 & 0.613
        & 0.807 & 0.807 & 0.810
        & 0.428 & 0.406 & 0.406 \\
\midrule
\multirow{3}{*}{MLP}
  & 42  & 0.468 & 0.518 & 0.525
        & 0.798 & 0.791 & 0.796
        & 0.472 & 0.498 & 0.510 \\
  & 123 & 0.472 & 0.503 & 0.511
        & 0.802 & 0.788 & 0.793
        & 0.461 & 0.478 & 0.489 \\
  & 456 & 0.455 & 0.487 & 0.494
        & 0.795 & 0.782 & 0.788
        & 0.467 & 0.491 & 0.503 \\
\bottomrule
\end{tabular}
\end{adjustbox}
\end{table}

\FloatBarrier
\section{Additional Details}\label{app:additional-details}

Table~\ref{tab:ablation} consolidates the numerical values behind the
ablation summaries of Figure~\ref{fig:ablation}.
The hyperparameter sensitivity sweep on HH-RLHF (seed~42, $1{,}000$ training
pairs) is visualized in Figure~\ref{fig:sensitivity}; the chosen
configuration is highlighted in that figure, and
Tables~\ref{tab:sensitivity-dim}, \ref{tab:sensitivity-l1},
and~\ref{tab:sensitivity-scale} list the per-cell numerical values
(all single-seed, as flagged in \S\ref{sec:ablation-studies}).
Each sweep is an independent single-seed execution; the nominally shared
configuration ($2K{=}16$, $\lambda_{\textup{reg}}{=}0.001$, $\alpha{=}0.5$)
therefore appears with slightly different values across panels
($\mathrm{acc}_\pi$ $0.515$ in Table~\ref{tab:sensitivity-dim} vs.\ $0.493$
in Tables~\ref{tab:sensitivity-l1}--\ref{tab:sensitivity-scale}), an
instance of the single-seed run-to-run variability flagged in
\S\ref{sec:ablation-studies}. Values are comparable within a panel but
should not be compared across panels.

\begin{table}[!t]
\centering
\caption{Ablation studies across three datasets. Panel~(a): component ablation
on $\mathrm{acc}_\pi$, progressively adding the curl head and $\ell_1$
regularization. Panel~(b): gradient coupling ablation on
$\mathrm{acc}_{\textup{dpo+c}}$, comparing SHPO with a stop-gradient
(Detach) variant. Panel~(c): inductive bias control on $\mathrm{acc}_\pi$,
comparing SHPO against an antisymmetrized MLP of matched parameter count.
DPO, HPO, SHPO, and Detach report 3-seed mean $\pm$ std (matching
Table~\ref{tab:main-results}); MLP values are 3-seed means.
\textbf{Bold}: best per dataset within each panel.}
\label{tab:ablation}
\small
\setlength{\tabcolsep}{5pt}
\begin{tabular}{l ccc}
\toprule
 & HH-RLHF & Synthetic & UltraFeedback \\
\midrule
\multicolumn{4}{l}{\textit{(a) Component ablation ($\mathrm{acc}_\pi$)}} \\
\midrule
DPO (baseline)
    & \sig{0.449}{0.049}
    & \sig{0.784}{0.017}
    & \sig{0.436}{0.009}
    \\
HPO (no $\ell_1$)
    & \sig{0.480}{0.016}
    & \sig{0.815}{0.007}
    & \sig{0.487}{0.049} \\
\textbf{SHPO}
    & \sig{\textbf{0.505}}{0.007}
    & \sig{\textbf{0.820}}{0.017}
    & \sig{\textbf{0.532}}{0.010} \\
\midrule
\multicolumn{4}{l}{\textit{(b) Gradient coupling --- Detach ablation ($\mathrm{acc}_{\textup{dpo+c}}$)}} \\
\midrule
\textbf{SHPO}
    & \sig{\textbf{0.609}}{0.010}
    & \sig{\textbf{0.821}}{0.020}
    & \sig{\textbf{0.656}}{0.014} \\
Detach (stop-grad)
    & \sig{0.508}{0.093}
    & \sig{0.730}{0.070}
    & \sig{0.572}{0.144} \\
$\Delta$
    & $+0.101$ & $+0.091$ & $+0.084$ \\
\midrule
\multicolumn{4}{l}{\textit{(c) Inductive bias --- MLP control ($\mathrm{acc}_\pi$)}} \\
\midrule
\textbf{SHPO}
    & $\mathbf{0.505}$ & $\mathbf{0.820}$ & $\mathbf{0.532}$ \\
MLP (matched params)
    & $0.465$ & $0.798$ & $0.467$ \\
$\Delta$
    & $+0.040$ & $+0.022$ & $+0.065$ \\
\bottomrule
\end{tabular}
\end{table}

\begin{table}[!t]
\centering
\caption{Curl dimension ($2K$) sweep on HH-RLHF (seed 42, $1{,}000$ training pairs).
\textbf{Bold}: best per metric column.}
\label{tab:sensitivity-dim}
\small
\setlength{\tabcolsep}{6pt}
\begin{tabular}{r ccc}
\toprule
$2K$ & $\mathrm{acc}_\pi$ & $\mathrm{acc}_{\textup{dpo}}$ & $\mathrm{acc}_{\textup{dpo+c}}$ \\
\midrule
4  & 0.492 & 0.550 & 0.562 \\
8  & 0.490 & 0.574 & 0.579 \\
\textbf{16} & \textbf{0.515} & \textbf{0.615} & \textbf{0.648} \\
32 & 0.434 & 0.447 & 0.453 \\
64 & 0.447 & 0.440 & 0.444 \\
\bottomrule
\end{tabular}
\end{table}

\begin{table}[!t]
\centering
\caption{$\ell_1$ regularization ($\lambda_{\textup{reg}}$) sweep on HH-RLHF (seed 42, $1{,}000$ training pairs).
\textbf{Bold}: best per metric column.}
\label{tab:sensitivity-l1}
\small
\setlength{\tabcolsep}{6pt}
\begin{tabular}{r ccc}
\toprule
$\lambda_{\textup{reg}}$ & $\mathrm{acc}_\pi$ & $\mathrm{acc}_{\textup{dpo}}$ & $\mathrm{acc}_{\textup{dpo+c}}$ \\
\midrule
0.0   & 0.448 & 0.462 & 0.462 \\
\textbf{0.001} & \textbf{0.493} & \textbf{0.594} & \textbf{0.604} \\
0.01  & 0.450 & 0.474 & 0.475 \\
0.05  & 0.487 & 0.555 & 0.560 \\
0.1   & 0.448 & 0.440 & 0.442 \\
\bottomrule
\end{tabular}
\end{table}

\begin{table}[!t]
\centering
\caption{Curl scale ($\alpha$) sweep on HH-RLHF (seed 42, $1{,}000$ training pairs).
\textbf{Bold}: best per metric column.}
\label{tab:sensitivity-scale}
\small
\setlength{\tabcolsep}{6pt}
\begin{tabular}{r ccc}
\toprule
$\alpha$ & $\mathrm{acc}_\pi$ & $\mathrm{acc}_{\textup{dpo}}$ & $\mathrm{acc}_{\textup{dpo+c}}$ \\
\midrule
0.1 & 0.443 & 0.469 & 0.470 \\
\textbf{0.5} & \textbf{0.493} & \textbf{0.588} & \textbf{0.595} \\
1.0 & 0.489 & 0.579 & 0.587 \\
2.0 & 0.446 & 0.476 & 0.484 \\
5.0 & 0.450 & 0.435 & 0.440 \\
\bottomrule
\end{tabular}
\end{table}

Table~\ref{tab:curl-diagnostics} reports diagnostics of the learned curl
component under the combinatorial Hodge decomposition
framework~\cite{jiang2011,lim2020}.
Table~\ref{tab:experiment-config} lists the full experimental
configuration. It covers the Qwen2.5 base model~\cite{qwen2024} and the
LoRA setup~\cite{hu2022lora} built on PEFT~\cite{peft2022} and
Accelerate~\cite{accelerate2022}.

\begin{table}[!t]
\centering
\caption{Curl head diagnostics. Panels~(a)--(b) are from the HH-RLHF checkpoint (1{,}000 training pairs, curl dimension $2K{=}16$,
epoch~$3$): (a)~sparsity statistics of the learned $\boldsymbol{\gamma}$;
(b)~curl contribution ratios on the HH-RLHF test set ($n{=}1000$). Panel~(c)
characterizes the cyclic structure of the Synthetic split ($794$ triangles): the prevalence of cyclic vs.\ transitive triangles (a data-level property) together with the curl head's direction agreement on the cyclic triangles.}
\label{tab:curl-diagnostics}
\small
\setlength{\tabcolsep}{6pt}
\begin{tabular}{l r}
\toprule
\multicolumn{2}{l}{\textit{(a) Sparsity statistics of learned $\boldsymbol{\gamma}$}} \\
\midrule
Number of curl planes ($K$)                       & $8$      \\
Active planes ($|\gamma_l| > 0.001$)              & $2$      \\
Active ratio                                       & $25.0\%$ \\
$\|\boldsymbol{\gamma}\|_1$ (all planes)           & $0.0705$ \\
$\|\boldsymbol{\gamma}\|_1$ (active planes)        & $0.069$  \\
\midrule
\multicolumn{2}{l}{\textit{(b) Curl contribution ratio (HH-RLHF test, $n{=}1000$)}} \\
\midrule
Mean ratio                                         & $18.13\%$ \\
Median ratio                                       & $7.90\%$  \\
P90 ratio                                          & $36.64\%$ \\
Fraction with ratio $> 50\%$                       & $8.73\%$  \\
\midrule
\multicolumn{2}{l}{\textit{(c) Cycle detection (Synthetic)}} \\
\midrule
Complete triangles analyzed                        & $794$                 \\
Cyclic triangles                                   & $169\ (21.28\%)$      \\
Transitive triangles                               & $625\ (78.72\%)$      \\
Direction agreement on cyclic triangles            & $76.4\%$              \\
\bottomrule
\end{tabular}
\end{table}

\begin{table}[!t]
\centering
\caption{Key experiment configuration (shared across all methods unless noted).
SHPO-specific hyperparameters were selected via grid search over HH-RLHF
(see hyperparameter sensitivity analysis in this appendix).}
\label{tab:experiment-config}
\small
\setlength{\tabcolsep}{6pt}
\begin{tabular}{l l}
\toprule
\multicolumn{2}{l}{\textit{(a) Training (shared across all methods)}} \\
\midrule
Base model                        & Qwen2.5-7B-Instruct                   \\
LoRA rank $r$ / $\alpha$          & $16$ / $32$                           \\
LoRA dropout                      & $0.05$                                \\
Quantization                      & QLoRA 4-bit (NF4)                     \\
Precision                         & bfloat16                              \\
Epochs                            & $3$                                   \\
Max sequence length               & $512$ tokens                          \\
Learning rate                     & $1{\times}10^{-4}$ (cosine schedule)  \\
Batch size (per-device)           & $1$                                   \\
Gradient accumulation             & $2$ steps                             \\
DPO temperature $\beta$           & $0.1$                                 \\
Random seeds                      & $\{42,\ 123,\ 456\}$                  \\
\midrule
\multicolumn{2}{l}{\textit{(b) SHPO-specific hyperparameters}} \\
\midrule
Curl planes $K$                   & $8$ (curl dimension $2K{=}16$)        \\
Curl scale $\alpha$               & $0.5$                                 \\
$\ell_1$ coefficient $\lambda_{\textup{reg}}$ & $0.001$                   \\
Detach curl                       & \texttt{false}                        \\
\bottomrule
\end{tabular}
\end{table}

\FloatBarrier

\newpage

\bibliographystyle{elsarticle-num}
{\sloppy
\bibliography{reference}}